\documentclass{article}

\usepackage[preprint]{corl_2026} 

\usepackage[utf8]{inputenc}
\usepackage[T1]{fontenc}
\usepackage{hyperref}
\usepackage{url}
\usepackage{booktabs}
\usepackage{amsfonts}
\usepackage{nicefrac}
\usepackage{microtype}
\usepackage{xcolor}
\usepackage{multicol}
\usepackage{multirow}
\usepackage{tikz}
\usepackage{pifont}
\usepackage{threeparttable}
\usepackage{makecell}
\usepackage{graphicx}
\usepackage{subcaption}
\usepackage[nointegrals]{wasysym}
\usepackage{amsmath,amssymb}
\usepackage{cleveref}
\usepackage{rotating}
\usepackage{longtable, booktabs}

\newcommand{\full}{\CIRCLE}       
\newcommand{\parcirc}{\LEFTcircle}
\newcommand{\none}{\Circle}       
\newcommand{\std}[1]{$\pm#1$}  

\title{CometVLA: Co-Training on an Embodied Data Pyramid towards Physical Understanding}

\author{Hanwen Wan$^{1, 2, 3}$, Dafeng Chi$^{1}$, Linbo Zhai$^{1, 5}$, Tianao Shen$^{2,3}$, \\ Yuzheng Zhuang$^{1}$, Tianle Zhang$^{1}$, Peidong Liu$^{1}$, Liang Lin$^{1, 6}$, and Xiaoqiang Ji$^{2,3,4,\dag}$%
\thanks{$^{1}$JD Explore Academy, China.}%
\thanks{$^{2}$School of Science and Engineering, The Chinese University of Hong Kong, Shenzhen, China.}%
\thanks{$^{3}$Shenzhen Institute of Artificial Intelligence and Robotics for Society, China.}%
\thanks{$^{4}$School of Artificial Intelligence, The Chinese University of Hong Kong, Shenzhen, China.}%
\thanks{$^{5}$South China University of Technology, China.}%
\thanks{$^{6}$Sun Yat-sen University, China}%
\thanks{$^{\dagger}$The corresponding author is Xiaoqiang Ji whose e-mail is {\tt\small jixiaoqiang@cuhk.edu.cn}}%
}

\begin{document}
\maketitle

\begin{figure}[h]
  \centering
  \includegraphics[width=\textwidth]{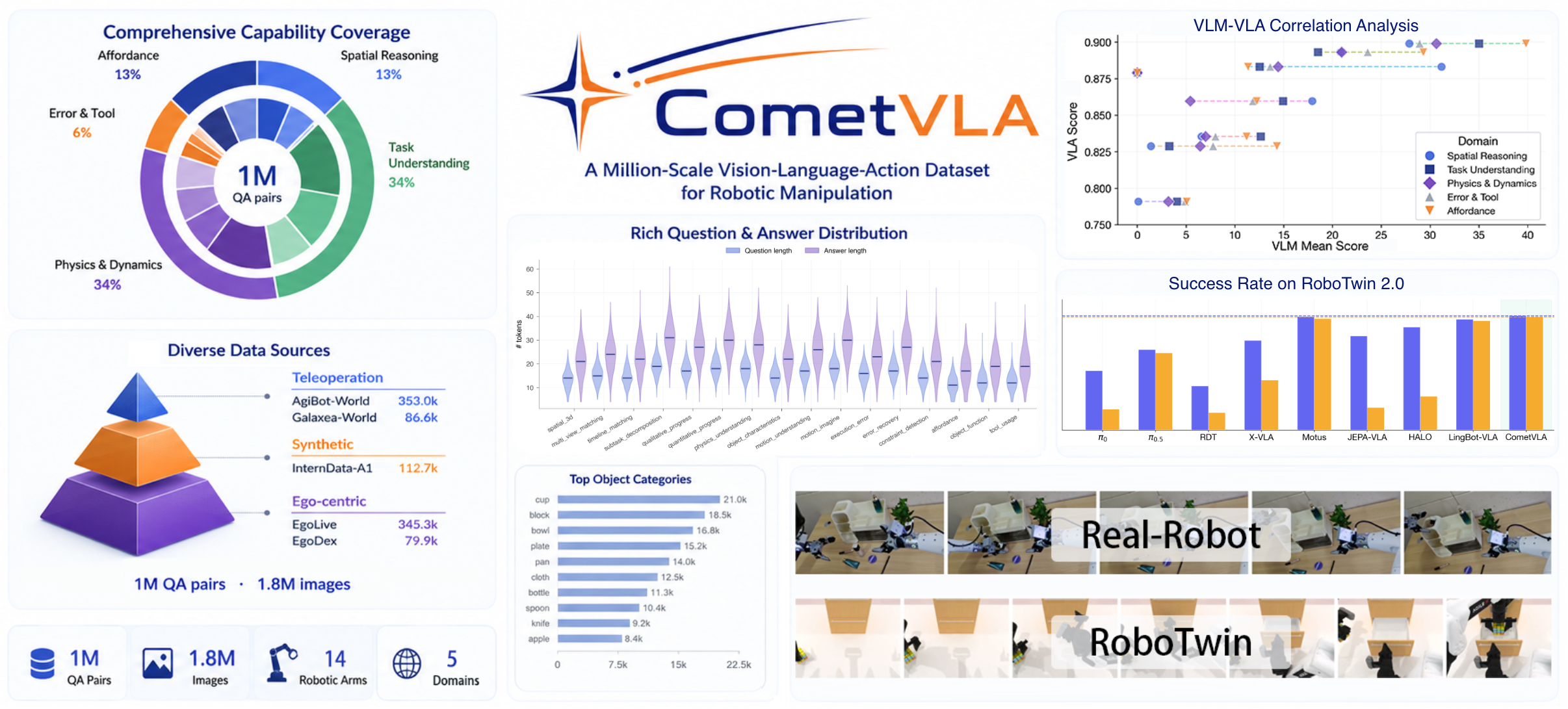}
  \caption{CometVLA overview.}
  \label{fig:teaser}
\end{figure}


\begin{abstract}
Vision-language-action (VLA) models remain brittle in manipulation tasks that require physical commonsense. Current physical VQA data is typically disembodied and misaligned with robot action domains. Egocentric videos are used only as auxiliary pre‑training. It remains unclear whether improved VLM physical understanding actually benefits downstream action generation. Therefore, we present CometVLA to close this gap. We construct CometData and CometBench, an embodied physical VQA corpus and benchmark strictly aligned with the robot’s action data and embodiment. We introduce Global Action Prior (GAP) tokens, a compact learnable bottleneck that isolates task‑agnostic motion regularities and lets the action head consume physical commonsense without corrupting the pre-trained VLM backbone. We co-train CometVLA across the embodied data pyramid, spanning teleoperation, simulation, egocentric trajectories, and VQA layers. On real‑world manipulation tasks and RoboTwin simulation, CometVLA consistently outperforms strong VLA baselines. Correlation analysis shows that stronger VLM performance on CometBench indicates higher VLA success rates. Results demonstrate that physical understanding pre-training genuinely benefits downstream manipulation.
\end{abstract}


\section{Introduction}
\label{sec:intro}

Vision-Language-Action models have made remarkable strides by repurposing large visuo-linguistic backbones as the perceptual and planning front-end of manipulation policies, inheriting grounding, spatial reasoning, and instruction following essentially for free~\cite{rt2, openvla, pi0}. VLM4VLA~\cite{vlm4vla} has shown that while VLM pre-training and VLA training share a broadly aligned learning trajectory early on, their representations eventually diverge into distinct regions. Although starting from fine-tuned VLMs~\cite{rt1, rt2}, they are systematically misaligned with the contact-rich, temporally evolving physical dynamics that are required by VLAs. This mismatch reveals a clear gap between VLMs and VLAs, and closing it is essential for physically reliable manipulation.

Physical commonsense is widely acknowledged as indispensable for robust robotic manipulation. Despite this recognition, existing efforts to incorporate physical knowledge into VLMs have largely remained confined to static settings~\cite{zhou}. Prior work typically injects physical priors through curated Visual Question Answering (VQA) datasets~\cite{causalvqa}, supervised fine-tuning~\cite{prismatic}, or alignment techniques~\cite{genrl}, and evaluates the resulting capabilities on disembodied benchmarks~\cite{phystoolbench}. While recent studies have begun to explore transferring such capabilities to embodied settings, they often decouple perception from action~\cite{hy,mimo} or simplify the control problem to high-level planning or discrete primitives~\cite{vlm4vla}, leaving open a fundamental question of whether improved physical understanding in VLMs actually translates to better action generation in realistic manipulation scenarios. From the perspective of data sources, most existing VQA corpora are harvested from internet images or simulation renderings~\cite{comprehensive}, while Robo2VLM~\cite{robo2vlm} attempts to extract VQA pairs directly from robot trajectories. Nevertheless, the VQA scenarios rarely match the data distribution of the robot action stream or the specific embodiment used during policy training in most cases, resulting in a persistent domain gap between physical reasoning and physical acting. 

Beyond VQA data, another important component of the embodied data pyramid is ego-centric human video, which provides scalable, diverse, and natural interaction data from real-world environments~\cite{innon}. Such videos are appealing because they capture fine-grained manipulation behaviors at scale with relatively low collection cost. Existing robot foundation models typically either treat ego-centric video as an auxiliary pre-training corpus mixed into the overall data blend~\cite{joyra01}, or leverage it to learn world models~\cite{fastwam} or latent representations~\cite{being} for downstream control policies. In contrast, our method introduces an alternative paradigm by curating physical-understanding VQA pairs from ego-centric video for co-training.

This analysis exposes two complementary gaps that explain why physical commonsense has so far failed to translate into reliable manipulation. We propose \textbf{CometVLA}, an end-to-end framework with physical understanding. We take a systematic route among data, model, and training. (i) At the data level, we build \textbf{CometData} and CometBench, an embodied physical-commonsense corpus and benchmark aligned with the recipe of the training robot data. Specifically, we curate CometData following the same recipe used for the VLA training dataset, thereby ensuring in-domain physical understanding enhancement. (ii) At the model architecture level, we introduce Global Action Prior (GAP) tokens, the information bottleneck for the action head to actively consume physical commonsense and to exploit motion regularities shared across various samples. (iii) At the training level, we implemented a single pre-training stage by leveraging knowledge insulation~\cite{ki}. The co-training paradigm couples all layers of the embodied data pyramid along with structural designs. The resulting system demonstrates robust action generation performance on both simulation benchmarks and real-world manipulation tasks. Moreover, the pre-trained models exhibit a positive correlation between performance on CometBench and action generation. Per-domain analysis reveals the contribution of different QA tasks.

In summary, the contributions of this work are: (1) CometData and CometBench (data). The first embodied physical-commonsense VQA corpus and benchmark that is aligned with the visual domain and embodiment of the robot action data used for policy training. Validation on CometBench shows a positive correlation with action generation. (2) Global Action Prior token (architecture). An inductive bias that lets the action head explicitly consume physical commonsense and aggregate cross-sample motion priors. (3) CometVLA (training). Training with embodied data pyramid for end-to-end action generation with physical understanding. Evaluations show robust real-world manipulation skills and superior performance compared to SOTA methods.

\section{Related Work}
\label{sec:related}

\subsection{Action Generation with Embodied Data Pyramid}

A fundamental challenge for embodied agents is the scarcity of high-quality, ego-centric robot data. To address this, recent work adopts an embodied data pyramid. Levels in the pyramid vary by source and collection method. The pyramid has a broad base of abundant, low-cost data and narrower upper layers of scarcer, more expensive data. From bottom to top: VQA data, ego-centric human videos~\cite{egodex,ego4d}, simulation data~\cite{interna1}, and finally real robot teleoperation~\cite{agibot,gal}. This pyramid offers a data-centric view of different action generation methods: imitation learning policies~\cite{ACT}, Vision-Language-Action models (VLAs), and World Action Models (WAMs)~\cite{fastwam}. Each leverages different combinations of pyramid levels to predict actions or their consequences.

Initial attempts typically consume the top, most expensive tier of the pyramid. RT-2~\cite{rt2} first showed that a web-scale VLM could be fine-tuned on a modest amount of teleoperation data to directly output action tokens. OpenVLA~\cite{openvla}, Octo~\cite{octo}, and $\pi_0$/$\pi_{0.5}$~\cite{pi0,pi05} further refined this paradigm, demonstrating that the language-conditioned action distribution can be effectively learned from hundreds of hours of real robot demonstrations. The pyramid's middle tier, consisting of simulation and scripted data, has been exploited by InternVLA\_A1~\cite{internvla_a1} to pre-train on human video, synthetic data, and real-world data. In contrast, WAMs aim to leverage the abundant ego-centric human video tier to train video generation abilities. EgoScale~\cite{egoscale} and FastWAM~\cite{fastwam} learn predictive world models from large-scale, unlabeled ego-centric video, enabling the agent to imagine future states without requiring action labels. Our work leverages each pyramid tier to form a comprehensive framework that systematically correlates embodied VLM capabilities with downstream VLA action performance, without conflating architectural changes or sacrificing action expert fidelity.

\subsection{Fine-tuning VLMs with Embodied/Physical VQAs}
\label{sec:related_phyvqa}
A growing amount of work fine-tunes VLMs on curated VQA corpora targeting embodied or physical reasoning~\cite{embodiedqa, openeqa, robovqa, erqa, physbench, physvlm, robobrain}, see~\cref{app:embodied_vqa} for a detailed comparison. Each introduces a dedicated fine-tuning dataset that emphasizes a distinct slice of the competence space, such as 2D/3D grounding, affordance reasoning, task planning, or error detection. Such models can be viewed as explicit carriers of capabilities to be inherited by downstream robot control. However, three limitations persist. First, most corpora remain disembodied, drawing their visual input from generic web imagery or short third-person videos rather than robot-centric observations. Second, although incorporating visual cues such as bounding boxes may provide guidance and improve performance, particularly for grounding and affordance tasks, such annotations are difficult to obtain and deploy in real-world settings. Third, few of these benchmarks have been validated in realistic VLA scenarios. Moreover, several methods rely on backbones of 32B parameters or larger, whose inference latency yields control frequencies too low to be practical for current VLA deployment. In contrast, our approach is sourced exclusively from robot action data spanning teleoperation, simulation, and ego-centric robot observations, ensuring direct alignment with real-world VLA deployment scenarios.

\subsection{Correlation Analysis from VLMs to VLAs}
\label{sec:related_vlm4vla}
Since RT-2~\cite{rt2} first repurposed a web-scale VLM as the backbone of a manipulation policy, a steady stream of vision-language-action (VLA) models, e.g., OpenVLA~\cite{openvla} and $\pi_0$~\cite{pi0}, has converged on a shared premise: general visuo-linguistic pre-training supplies the grounding and instruction following that scarce robot data can no longer afford to teach from scratch. However, systematic investigation of how action generation benefits from VLMs remains absent. Recent VLA works have leveraged embodied VQA data while simultaneously improving multiple components, including architecture, data, and training strategies, to boost VLA performance, as seen in MIMO-Embodied~\cite{mimo}, HY-Embodied-0.5~\cite{hy}, and GenieReasoner~\cite{liu2025unified}. While effective, these approaches couple various improvements, making it difficult to isolate the specific contribution of VLM representations. In contrast, VLM4VLA~\cite{vlm4vla} has taken a first step toward correlating VLMs and VLAs by simplifying the action expert to a minimal set of parameters. While this successfully isolates the influence of VLM representations from different action expert designs, it also limits the practical applicability of its conclusions, as the action expert network is an inherent and non-separable variable in the VLA pipeline. Complementing this, Lin et al.~\cite{lin2026} provide large-scale evidence that the VLM backbone's understanding directly correlates with policy generalization even under a full, non-simplified action expert, showing that effective co-training preserves VLM capabilities and yields stronger robot performance. In contrast to these existing efforts, we proposed a different analytical framework that neither couples all components nor simplifies the action expert. This allows a systematic assessment of how specific VLM capacities translate into downstream action performance.


\begin{figure*}[t]
    \centering
    \includegraphics[width=\textwidth]{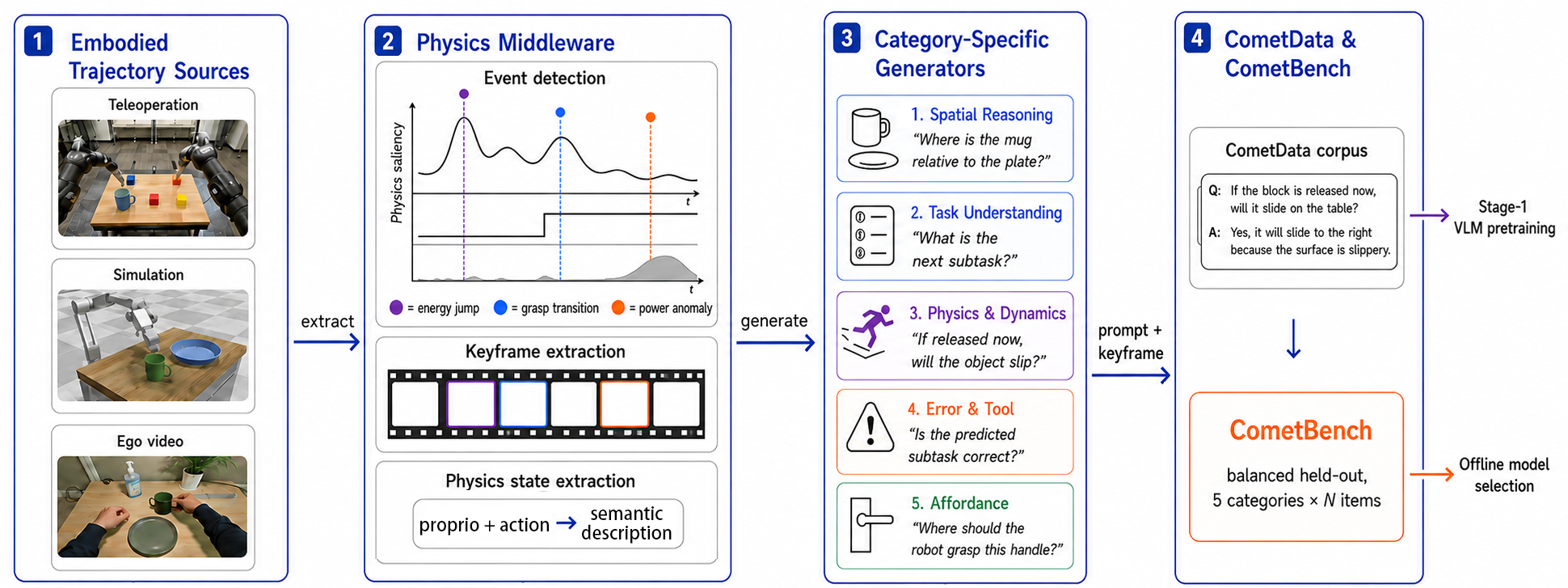}
    \caption{
    Overview of the proposed CometData generation pipeline for physics-aware embodied VLMs. (1) Multi-source embodied trajectories are collected from teleoperation, simulation, and ego-centric videos, forming a hierarchical embodied data pyramid with increasing physical grounding. (2) A physics middleware extracts keyframes, detects interaction events and converts low-level proprioceptive and action signals into semantic physical state descriptions. (3) Category-specific generators construct diverse embodied VQA tasks, spanning 5 domains with 16 categories. (4) The generated samples are curated into the CometData corpus and the sampled CometBench benchmark.
    }
    \label{fig:comet_pipeline}
\end{figure*}

\section{Method}
\label{sec:method}

\subsection{Overview}
\label{sec:method_overview}
CometVLA integrates physical commonsense into a VLA through a coherent sequence of design choices. We construct CometData, a large corpus of embodied physical VQA pairs automatically harvested from the full embodied data pyramid. We then leverage knowledge insulation to implement VLM-VLA co-pre-training as the first stage. This stage simultaneously trains the VLM backbone with a FAST action tokenizer and a flow-matching action expert. The pre-trained backbone is extended with a small set of learnable GAP tokens. These GAP tokens are globally shared and decoupled from the vision-language stream, forming a bottleneck that encodes task-agnostic motion regularities.

\subsection{CometData Data Construction}
\label{sec:method_data}

CometData is constructed through a physics-aware embodied data generation pipeline with 1 million QA pairs and incorporates more than 1.8 million images. We build a data pipeline that integrates heterogeneous trajectory sources, structured physical state extraction, and category-specific QA synthesis, as shown in~\cref{fig:comet_pipeline}. At the generator level, 16 generators from 5 domains transform these descriptors into question-answer pairs. See~\cref{app:generator} for details of the generators. Generators fill category templates with the middleware descriptors, renders the selected keyframes, and prompts a strong teacher VLM to produce grounded answers. Notably, we construct CometData following the same recipe of robot data for pre-training, which ensures an in-domain semantic enhancement.

At the core of CometData is a physics-aware middleware that consumes synchronized proprioceptive state \(s_t\) and commanded action \(a_t\), and emits compact descriptors of trajectory dynamics. Instead of uniform sampling, we identify manipulation‑critical keyframes via dynamics‑based criteria.~\cite{frameskip}. Keyframes are scored using three embodiment-agnostic criteria: Hamiltonian jump $\mathcal{C}_{\text{A}}(t)$, gripper step $\mathcal{C}_{\text{B}}(t)$, and power anomaly $\mathcal{C}_{\text{C}}(t)$. The saliency score $\mathcal{S}_t $ aggregates these indicators to prioritize physically salient frames. Keyframes are obtained by retaining the top-\(K\) frames per episode. See~\cref{app:keyframe} for more details.

\paragraph{CometBench}
A balanced held-out subset of CometData is reserved as CometBench. It comprises 2,000 questions with reference answers for evaluating embodied physical reasoning in vision-language models. CometBench tests co-trained VLAs using curated question-answer pairs that span five distinct domains from CometData. The evaluation is conducted by a large language model as a consistent and fair judge. For each question, the judge compares the candidate answer against the reference answer on correctness, logical coherence, relevance, completeness, and physical reasoning with a total of 100 scores. Results are averaged among the full 2000 questions of CometBench. All the QA-pairs are reviewed by human testers for integrity and soundness.

\subsection{CometVLA Architecture}
\label{sec:method_arch}

\begin{figure*}[t]
\centering
\includegraphics[width=\textwidth]{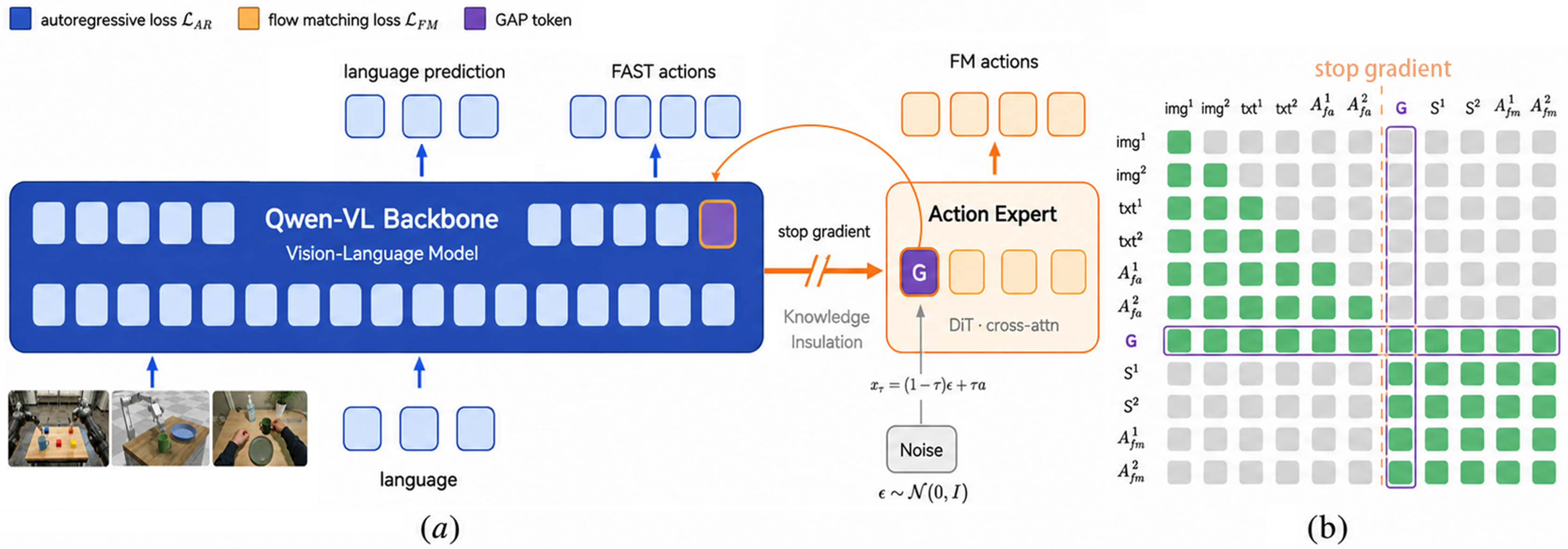}
\caption{
\textbf{CometVLA architecture and global action prior token.} \textbf{(a)} CometVLA unifies auto-regressive vision-language modeling and continuous action generation within a single framework. Multi-view observations, language tokens, and FAST actions are processed by a pre-trained Qwen3-VL backbone under the auto-regressive objective $\mathcal{L}_{\mathrm{AR}}$. A lightweight diffusion-based action expert predicts continuous FM actions using flow matching $\mathcal{L}_{\mathrm{FM}}$, while a stop-gradient barrier preserves pre-trained multimodal representations during policy learning. \textbf{(b)} Attention visibility grid of CometVLA. The orange dashed line denotes the stop-gradient barrier, and the purple block highlights the GAP token, which serves as the sole communication interface between the auto-regressive backbone and the diffusion action expert. $\text{img}^i$, $\text{text}^i$, $\mathrm{A}_{fa}^i$, $G$, $S^i$, and $\mathrm{A}_{fm}^i$ denote image tokens, language tokens, FAST action tokens, the GAP token, robot state tokens, and flow-matching action tokens, respectively.
}
\label{fig:arch}
\end{figure*}

CometVLA combines auto-regressive vision-language modeling with diffusion-based continuous action generation in a unified architecture, as illustrated in~\cref{fig:arch}. Given multi-view visual observations, proprioceptive states, and language instructions, a pre-trained Qwen3-VL~\cite{Qwen3-VL} backbone jointly models vision-language tokens and discrete FAST action tokens~\cite{fast} under a standard autoregressive objective $\mathcal{L}_{\mathrm{AR}} + \mathcal{L}_{\mathrm{fast}}$. Specifically, $\mathcal{L}_{\mathrm{AR}}$ is a cross-entropy loss for language generation, while $\mathcal{L}_{\mathrm{fast}}$ supervises discretized action tokens produced by the FAST tokenizer.

To enable high-frequency continuous control, we further attach a lightweight diffusion-based action expert trained with a flow-matching objective $\mathcal{L}_{\mathrm{fm}}$. Continuous actions are noised via a Beta-sampled timestep and optimized through velocity regression, where the model predicts the ground-truth flow field from latent action representations conditioned on VLM embeddings, GAP tokens, and state tokens. During training, a stop-gradient barrier isolates the action expert from the backbone, ensuring that $\mathcal{L}_{\mathrm{fm}}$ only updates the action head and GAP token parameters while preserving the pre-trained multimodal reasoning capability of the VLM. The overall training objective is:
\begin{equation}
\mathcal{L}_{\text{total}} = \mathcal{L}_{\text{AR}} + \mathcal{L}_{\text{fast}} + \mathcal{L}_{\text{fm}}.
\end{equation}

\paragraph{Global Action Prior}
To bridge the auto-regressive VLM domain and the bidirectional diffusion policy domain, we introduce a dedicated GAP token as the sole communication interface between the two modules. The idea of an information bottleneck has been widely validated for improving efficiency and modularity in VLMs~\cite{flamingo}. Recent VLA works have explored latent bottlenecks and compressed interfaces to decouple high-level reasoning from low-level control, such as latent intent variables in DIAL~\cite{dial} and token compression or latent action representations in VLA pre-training frameworks~\cite{lapa, compressorvla}. However, existing approaches typically rely on either fully shared representations or symmetric cross-attention, which can still lead to optimization interference between language modeling and low-level control.

In contrast, the GAP token is inserted directly into the backbone token sequence and transferred to the action expert through a detached latent pathway. Functionally, it serves as a global action prior that aggregates high-level semantic and temporal context within the VLM backbone, similar to a \texttt{[CLS]} token~\cite{bert}, and conditions action generation inside the DiT-based action expert via cross-attention. During attention computation, the GAP token is permitted to traverse the stop-gradient boundary, enabling controlled semantic information flow while preventing gradient leakage into the backbone. This asymmetric attention topology decouples language modeling from continuous control, improving optimization stability while preserving the pretrained multimodal reasoning capabilities.


\section{Experiments}
\label{sec:experiments}

\subsection{Experimental Setup}
\label{sec:exp_setup}

\paragraph{Training details}
We train CometVLA in two stages. In the co-training stage, we combine a Qwen3-VL-4B backbone, a FAST action tokenizer, and a DiT-B flow-matching action head with a learnable GAP token. The model is jointly trained on AgiBot-World-Beta~\cite{agibot}, InternData-A1~\cite{interna1}, EgoLive~\cite{joyra01}, EgoDex~\cite{egodex}, common VQA dataset\cite{cambrian} and the CometData VQA dataset. All robot datasets are retargeted to a bi-manual format with the action space padded to $32$ dimensions. We train distributed across $4$ nodes with $8\times$H200 GPUs each. In the post-training stage, we further fine-tune the model on task-specific datasets, including real-world teleoperation data for deployment and RoboTwin for simulation benchmarks, respectively.

\begin{figure}[t]
    \centering
    \setlength{\tabcolsep}{1pt}
    \renewcommand{\arraystretch}{0.2}
    \begin{tabular}{ccccc}
        \includegraphics[width=0.19\textwidth]{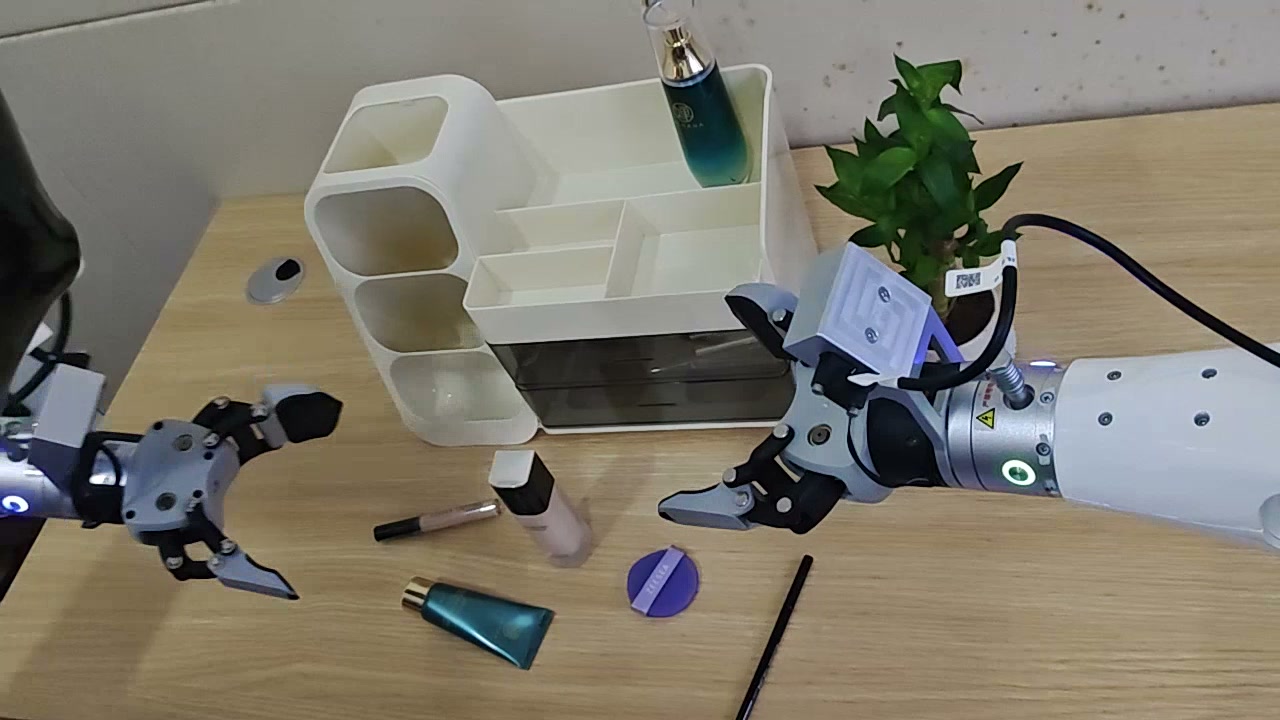} &
        \includegraphics[width=0.19\textwidth]{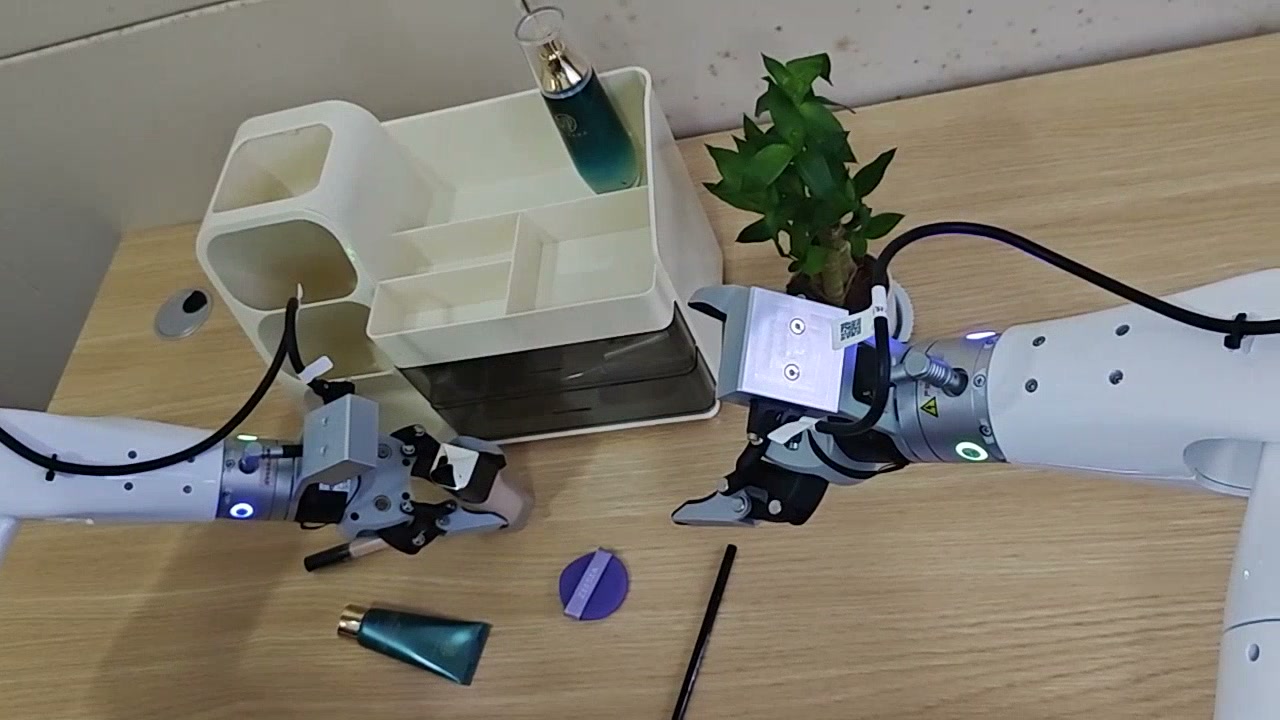} &
        \includegraphics[width=0.19\textwidth]{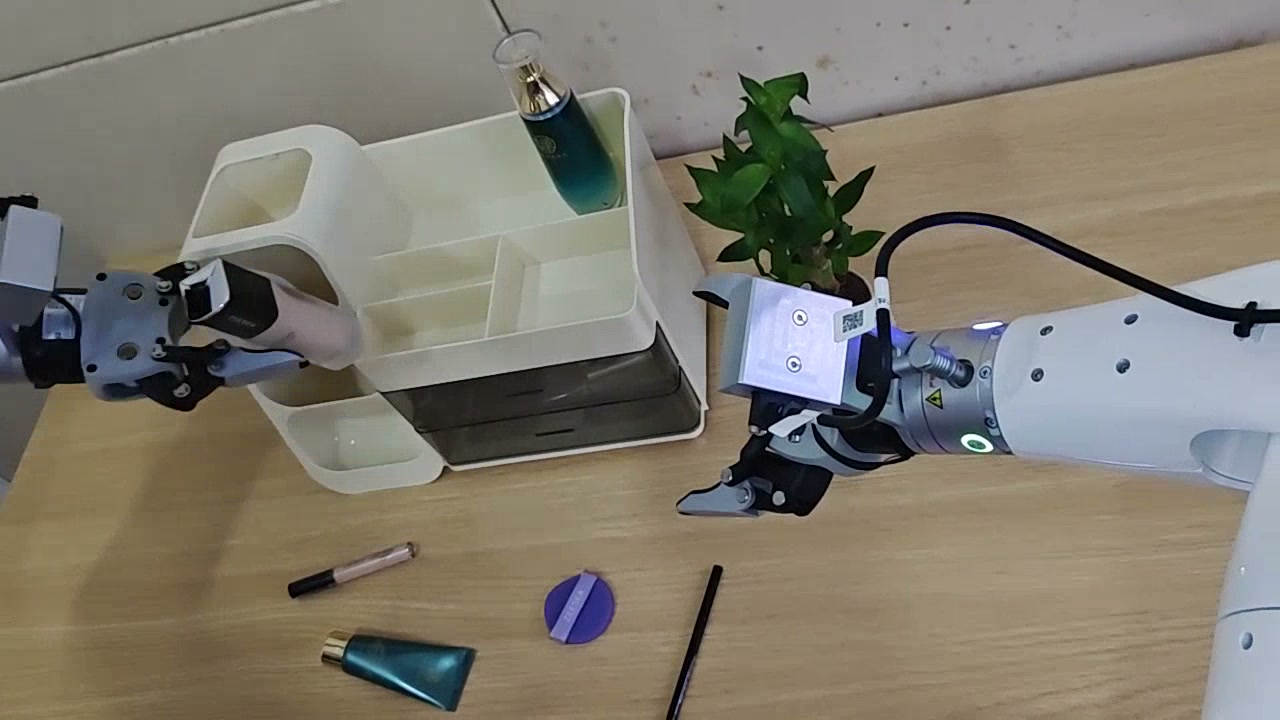} &
        \includegraphics[width=0.19\textwidth]{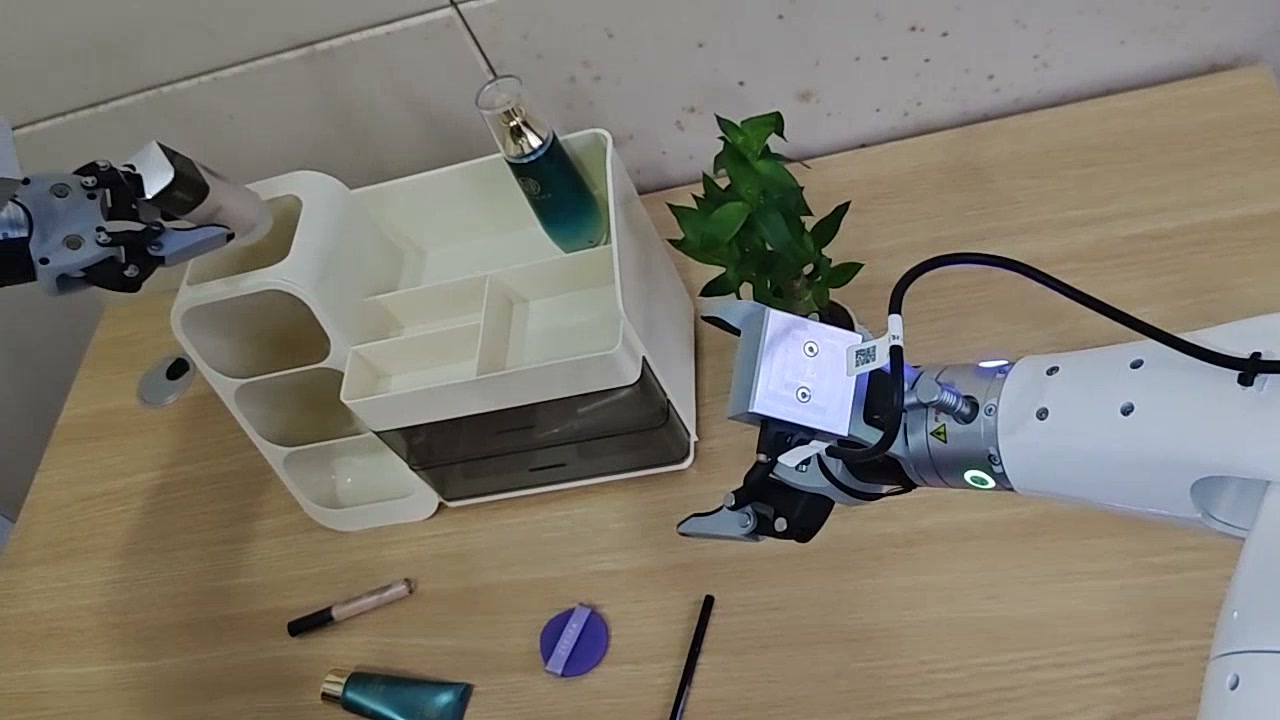} &
        \includegraphics[width=0.19\textwidth]{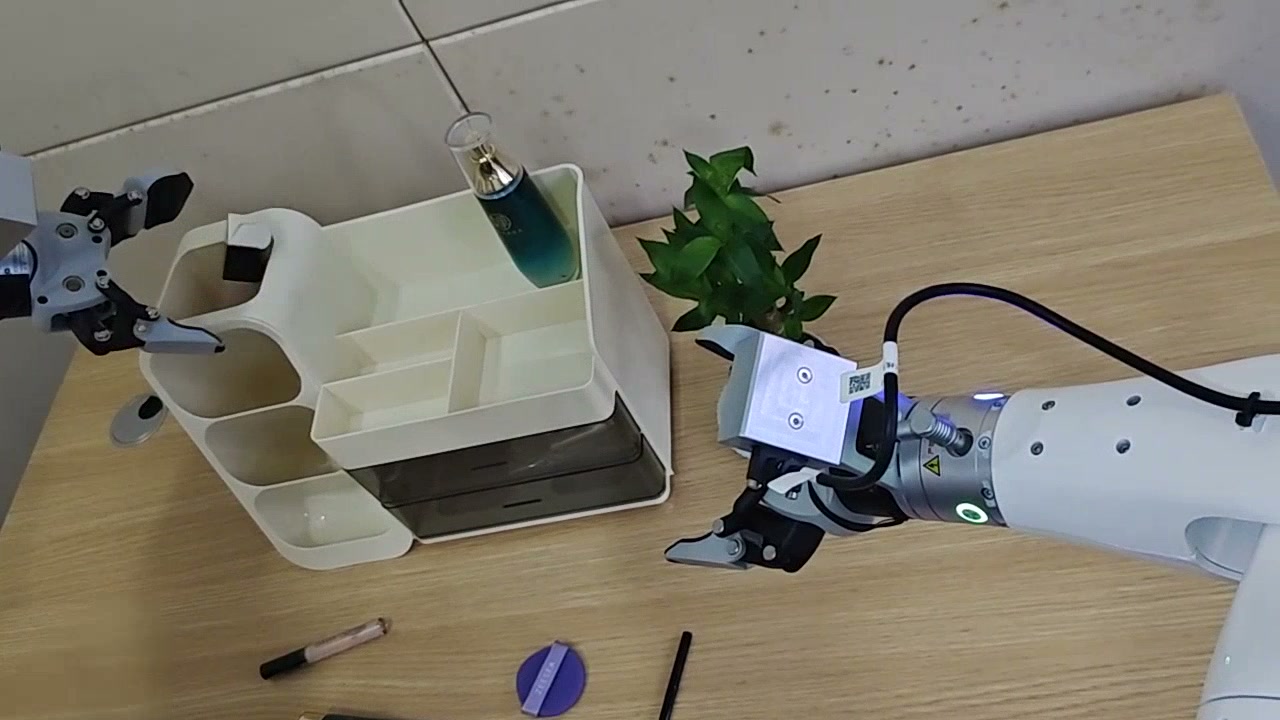}
    \end{tabular}
    \caption{Captured frames during the real-robot experiment of organizing the liquid foundation.}
    \label{fig:rollout_real}
\end{figure}

\begin{table}[t]
\centering
\caption{Success rate of real-robot experiment.}
\label{tab:real}
\begin{tabular}{lccccc}
\toprule
Task name & bb cream & liquid foundation & lotion & makeup sponge & serum \\
\midrule
SR (\%) & 75.00 & 31.25 & 37.50 & 93.75 & 43.75 \\
\bottomrule
\end{tabular}
\end{table}

\subsection{Real-Robot Experiments}
\label{sec:exp_realrobot}

We conduct real-world experiments on the G1 dual-arm robot from AgiBot Robotics. The manipulation policy runs on local workstation, forming the control client. We set up a dressing table organization scenario with multiple tasks. Pre-trained CometVLA is fine-tuned on task-specific datasets. We tested each task for 16 trials, results are recorded in~\cref{tab:real} and~\cref{fig:rollout_real}. Success rate varies across different objects resulting from object size and textures. More details are shown in~\cref{app:real}.

\subsection{Evaluation on RoboTwin}
\label{sec:exp_robotwin}

We compare CometVLA against representative VLA baselines spanning discrete-token, diffusion, and flow-matching action heads. As summarized in~\cref{tab:main_results}, CometVLA achieves an 89.24\% success rate on the RoboTwin easy (clean) split and 88.38\% on the hard (randomized) split, outperforming prior methods including $\pi_{0.5}$~\cite{pi05}. Ablations trained under the same configuration further validate the contribution of each proposed component. Replacing the curated embodied physical VQA corpus with same amount of generic VQA datasets (\textit{w/o phys. VQA}) during pre-training leads to the largest performance drop, indicating that physical understanding learned from embodied data substantially improves downstream manipulation performance. Removing the Global Action Prior (GAP) token also consistently degrades performance. Additional results are provided in the appendix, including per-task success rates in~\cref{app:rt-pertask}, qualitative rollout visualizations in~\cref{app:rt-sim}, and action trajectory analysis in~\cref{app:rt-action}.

\begin{table}[t]
\centering
\caption{\textbf{Comparison of VLA methods on RoboTwin 2.0.} Metrics originally reported with single-decimal precision are zero-padded.}
\label{tab:main_results}
\setlength{\tabcolsep}{4pt}
\resizebox{\linewidth}{!}{
\begin{tabular}{l c ccccccccccc}
\toprule

\multirow{2}{*}{Model} & \multirow{2}{*}{Split} & \multicolumn{8}{c}{Baseline Methods} & \multicolumn{3}{c}{Ours} \\
\cmidrule(lr){3-10}
\cmidrule(lr){11-13}
& & $\pi_0$ & $\pi_{0.5}$ & RDT & X-VLA & Motus & \makecell{JEPA\\-VLA} & HALO & \makecell{LingBot\\-VLA} & \makecell{w/o phys.\\VQA} & \makecell{w/o \\GAP} & \textbf{CometVLA} \\
\midrule
\multirow{2}{*}{RoboTwin} & Easy & 46.42 & 62.86 & 34.50 & 70.00 & 88.66 & 73.50 & 80.50 & 86.50 & 83.54 & 86.38 & \textbf{89.24} \\
& Hard & 16.34 & 60.30 & 13.72 & 39.00 & 87.02 & 17.70 & 26.40 & 85.34 & 83.46 & 85.12 & \textbf{88.38} \\
\bottomrule
\end{tabular}
}
\end{table}

\subsection{VLM-VLA Performance Correlation Analysis}
\label{sec:cometbench}

\begin{figure}[t]
    \centering
    \begin{subfigure}[c]{0.52\linewidth}
        \centering
        \includegraphics[width=\linewidth]{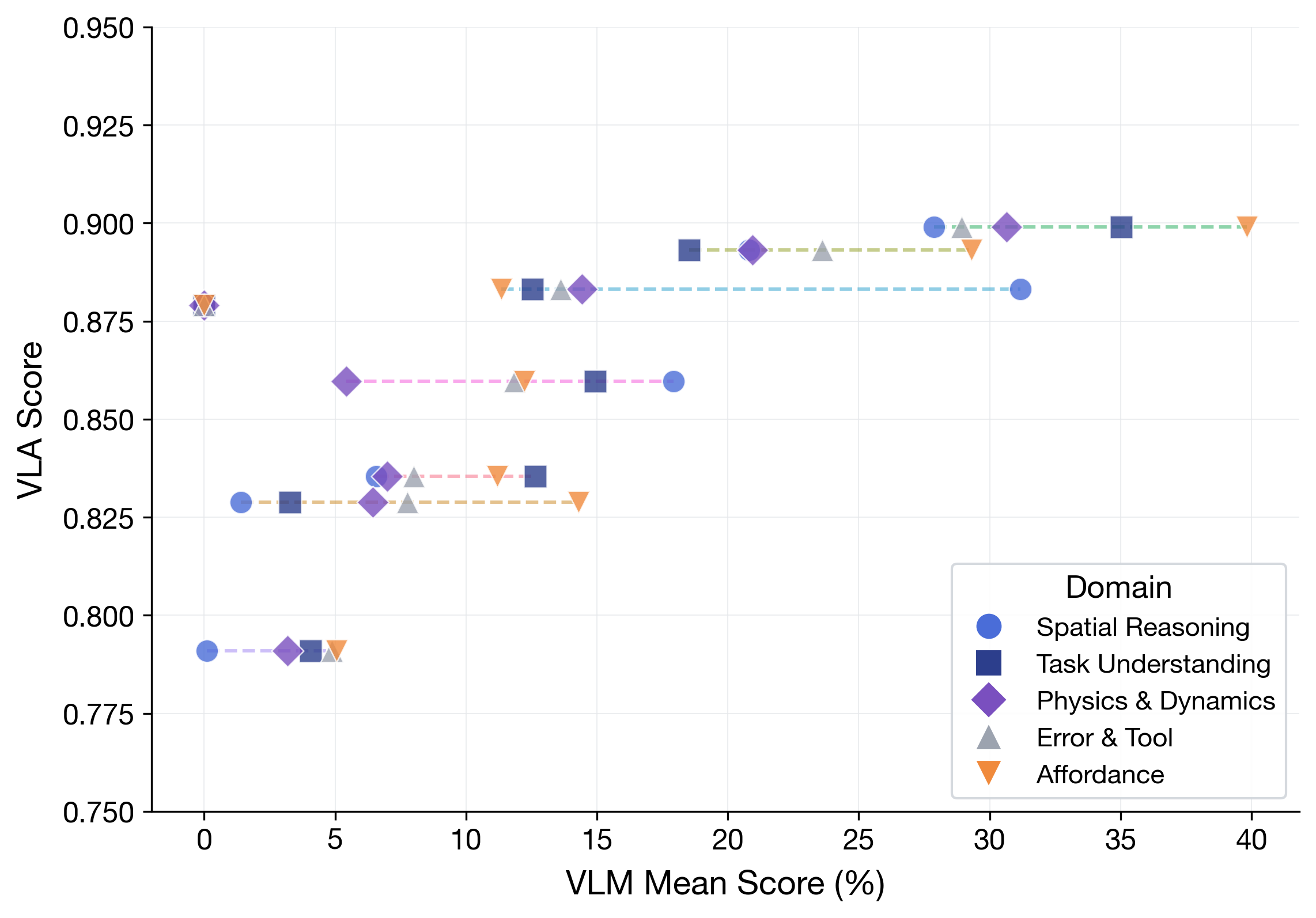}
        \caption{Visualization of the correlation analysis on CometBench and RoboTwin 2.0.}
        \label{fig:corr}
    \end{subfigure}
    \hfill
    \begin{subfigure}[c]{0.42\linewidth}
        \centering
        \includegraphics[width=\linewidth]{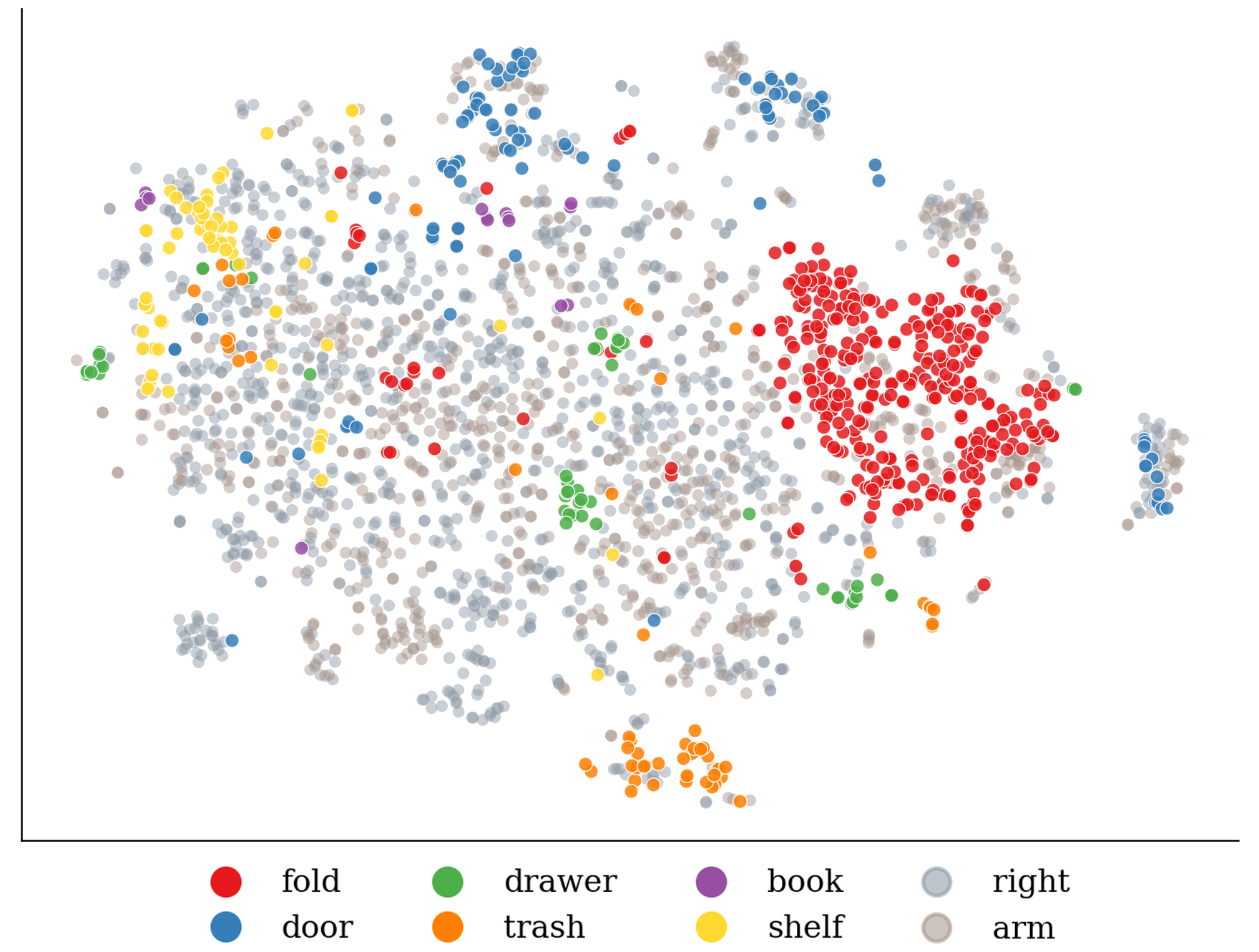} 
        \caption{t-SNE visualization of the GAP token on a held-out subset of AgiBot-World-Beta, colored by keyword clusters.}
        \label{fig:tsne}
    \end{subfigure}

    \vspace{4pt}

    \begin{subtable}[c]{0.41\linewidth}
        \centering
        \caption{Domain-wise Pearson correlation between VLM performance and VLA success rate.}
        \label{tab:domain_corr}
        \resizebox{\linewidth}{!}{%
        \begin{tabular}{lcc}
            \toprule
            Domain & Pearson $r$ & Count \\
            \midrule
            Spatial Reasoning   & 0.721 & 245 \\
            Task Understanding  & 0.586 & 326 \\
            Physics \& Dynamics & 0.645 & 413 \\
            Error \& Tool       & 0.624 & 601 \\
            Affordance          & 0.552 & 415 \\
            \bottomrule
        \end{tabular}
        }
    \end{subtable}
    \hfill
    \begin{subtable}[c]{0.56\linewidth}
        \centering
        \caption{Linear probing accuracy (\%) on action-relevant labels (5-fold CV).}
        \label{tab:action_probing}
        \resizebox{\linewidth}{!}{%
        \begin{tabular}{lccc}
            \toprule
            Task & Chance & VLM-mean & GAP (Ours) \\
            \midrule
            Gripper state    & 30.7 & 42.8\std{1.6} & \textbf{49.8\std{1.1}} \\
            Motion direction & 21.1 & 32.2\std{1.1} & \textbf{32.9\std{0.9}} \\
            Motion magnitude & 33.4 & 50.4\std{1.4} & \textbf{54.3\std{0.9}} \\
            \bottomrule
        \end{tabular}
        }
    \end{subtable}

    \caption{Analysis of VLM–VLA correlation, GAP token semantics, and action probing test.}
    \label{fig:gap_joint}
\end{figure}

We conduct a correlation analysis between VLM performance on CometBench and VLA performance on the RoboTwin 2.0 easy benchmark. See per-domain analysis in~\cref{app:percorr}. All results are obtained from a set of models trained under identical VLM backbone and action expert configurations with full parameters activated for training. These trials differ in robot action dataset mixtures or training configurations, yielding a diverse range of data points. As shown in~\cref{fig:corr} and summarized in~\cref{tab:domain_corr}, we observe a consistent positive correlation between VLM and VLA performance across all semantic domains. The overall trend indicates that stronger physical understanding is predictive of downstream robotic manipulation success. Among different categories, Spatial Reasoning exhibits the strongest correlation, suggesting that geometric and spatial priors transfer most effectively to embodied control. Physics \& Dynamics shows the second strongest correlation, which validates our proposed claim. Moreover, Task Understanding's moderate correlation may stem from the sub-task structure embedded in the robot action data, where additional enhancement is less critical.

\subsection{Analysis on GAP token}
\label{sec:exp_ablation}

A natural question is whether the bottleneck imposed by a single $1024$-d GAP token learns the desired physical commonsense aggregation of the $2560$-d VLM backbone. To examine this, we conduct two complementary analysis on a held-out subset of AgiBot-World-Beta. The first is a qualitative semantic preservation test via t-SNE and TF-IDF of token-level features. The second is a quantitative action information content test via linear probing on action-relevant labels. We compare the GAP token against a mean-pooled final-layer VLM representation, which serves as an unrestricted, higher-capacity baseline. See implementation details in App.~\cref{app:gap}.

We clustered task annotations with TF-IDF and visualize the GAP features with t-SNE, as shown in~\cref{fig:tsne}. The GAP token yields a t-SNE layout where broad embodiment concepts like “arm” and “left” form continuous global regions, and interaction-centric semantics like “bottle” and “door” emerge as localized clusters. This indicates that the bottleneck learns a semantic-level representation of the actions. Beyond semantics,~\cref{tab:action_probing} shows that the GAP token consistently outperforms VLM-mean on action-relevant probing. It achieves $+7.0\%$ on gripper state and $+3.9\%$ on motion magnitude. Taken together, these results support our central design claim that despite a $2.5\times$ smaller dimensionality, the GAP token preserves the backbone's semantic structure while selectively distilling control-relevant information into a compact latent interface for the action expert.


\section{Limitations}

Several limitations of the current study should be acknowledged. First, the correlation analysis may benefit from a wider range of data points. The corresponding VLA success rates are concentrated within a moderate range around 85\%. A more diverse set of model checkpoints with greater performance disparity would further strengthen the statistical basis of the observed correlation. Second, the real-robot evaluation could be extended with more scenarios and tasks. The current experiments cover only a subset of manipulation behaviors. Expanding to a broader set of environments and task difficulties would provide additional evidence for the generalizability of our findings. 

\section{Conclusion}
\label{sec:conclusion}

In this work, we presented CometVLA, an end-to-end vision-language-action framework for physical commonsense reasoning. We constructed CometData and CometBench—the first embodied physical VQA corpus and benchmark aligned with the robot action data pyramid, demonstrating positive correlation with downstream action performance. We introduced GAP tokens, a lightweight information bottleneck that enables the action head to explicitly consume physical commonsense and aggregate cross-sample motion regularities. We trained CometVLA with a co-training paradigm that jointly leverages the full embodied data pyramid. Extensive evaluations show that CometVLA achieves superior performance on both simulation benchmarks and real-world dual-arm manipulation tasks. Correlation analysis and action probing further validate the proposed methods.


\clearpage


\bibliography{example}  

@article{robopoint,
  title={Robopoint: A vision-language model for spatial affordance prediction for robotics},
  author={Yuan, Wentao and Duan, Jiafei and Blukis, Valts and Pumacay, Wilbert and Krishna, Ranjay and Murali, Adithyavairavan and Mousavian, Arsalan and Fox, Dieter},
  journal={arXiv preprint arXiv:2406.10721},
  year={2024}
}

@article{refspatial,
  title={Roborefer: Towards spatial referring with reasoning in vision-language models for robotics},
  author={Zhou, Enshen and An, Jingkun and Chi, Cheng and Han, Yi and Rong, Shanyu and Zhang, Chi and Wang, Pengwei and Wang, Zhongyuan and Huang, Tiejun and Sheng, Lu and others},
  journal={Advances in Neural Information Processing Systems},
  volume={38},
  pages={28404--28481},
  year={2026}
}

@article{robo2vlm,
  title={Robo2vlm: Visual question answering from large-scale in-the-wild robot manipulation datasets},
  author={Chen, Kaiyuan and Xie, Shuangyu and Ma, Zehan and Sanketi, Pannag R and Goldberg, Ken},
  journal={arXiv preprint arXiv:2505.15517},
  year={2025}
}

@article{cosmosreason1,
  title={Cosmos-reason1: From physical common sense to embodied reasoning},
  author={Azzolini, Alisson and Bai, Junjie and Brandon, Hannah and Cao, Jiaxin and Chattopadhyay, Prithvijit and Chen, Huayu and Chu, Jinju and Cui, Yin and Diamond, Jenna and Ding, Yifan and others},
  journal={arXiv preprint arXiv:2503.15558},
  year={2025}
}

@article{multiplan,
  title={Toward universal embodied planning in scalable heterogeneous field robots collaboration and control},
  author={Wan, Hanwen and Zhang, Yuhan and Wang, Junjie and Wu, Donghao and Li, Mengkang and Chen, Xilun and Deng, Yixuan and Huang, Yuxuan and Sun, Zhenglong and Zhang, Lin and others},
  journal={Journal of Field Robotics},
  volume={42},
  number={5},
  pages={2318--2336},
  year={2025},
  publisher={Wiley Online Library}
}

@misc{dial,
      title={DIAL: Decoupling Intent and Action via Latent World Modeling for End-to-End VLA}, 
      author={Yi Chen and Yuying Ge and Hui Zhou and Mingyu Ding and Yixiao Ge and Xihui Liu},
      year={2026},
      eprint={2603.29844},
      archivePrefix={arXiv},
      primaryClass={cs.RO},
      url={https://arxiv.org/abs/2603.29844}, 
}

@misc{flamingo,
      title={Flamingo: a Visual Language Model for Few-Shot Learning}, 
      author={Jean-Baptiste Alayrac and Jeff Donahue and Pauline Luc and Antoine Miech and Iain Barr and Yana Hasson and Karel Lenc and Arthur Mensch and Katie Millican and Malcolm Reynolds and Roman Ring and Eliza Rutherford and Serkan Cabi and Tengda Han and Zhitao Gong and Sina Samangooei and Marianne Monteiro and Jacob Menick and Sebastian Borgeaud and Andrew Brock and Aida Nematzadeh and Sahand Sharifzadeh and Mikolaj Binkowski and Ricardo Barreira and Oriol Vinyals and Andrew Zisserman and Karen Simonyan},
      year={2022},
      eprint={2204.14198},
      archivePrefix={arXiv},
      primaryClass={cs.CV},
      url={https://arxiv.org/abs/2204.14198}, 
}

@article{lapa,
  title={Latent Action Pretraining from Videos},
  author={Ye, Seonghyeon and Jang, Joel and Jeon, Byeongguk and Joo, Sejune and Yang, Jianwei and Peng, Baolin and Mandlekar, Ajay and Tan, Reuben and Chao, Yu-Wei and Lin, Bill Yuchen and others},
  journal={arXiv preprint arXiv:2410.11758},
  year={2024}
}

@misc{compressorvla,
      title={Compressor-VLA: Instruction-Guided Visual Token Compression for Efficient Robotic Manipulation}, 
      author={Juntao Gao and Feiyang Ye and Jing Zhang and Wenjing Qian},
      year={2025},
      eprint={2511.18950},
      archivePrefix={arXiv},
      primaryClass={cs.RO},
      url={https://arxiv.org/abs/2511.18950}, 
}

@misc{frameskip,
      title={FrameSkip: Learning from Fewer but More Informative Frames in VLA Training}, 
      author={Bin Yu and Shijie Lian and Xiaopeng Lin and Zhaolong Shen and Yuliang Wei and Changti Wu and Hang Yuan and Haishan Liu and Bailing Wang and Cong Huang and Kai Chen},
      year={2026},
      eprint={2605.13757},
      archivePrefix={arXiv},
      primaryClass={cs.RO},
      url={https://arxiv.org/abs/2605.13757}, 
}

@inproceedings{robospatial,
  title={Robospatial: Teaching spatial understanding to 2d and 3d vision-language models for robotics},
  author={Song, Chan Hee and Blukis, Valts and Tremblay, Jonathan and Tyree, Stephen and Su, Yu and Birchfield, Stan},
  booktitle={Proceedings of the Computer Vision and Pattern Recognition Conference},
  pages={15768--15780},
  year={2025}
}

@article{robobench,
  title={Robobench: A comprehensive evaluation benchmark for multimodal large language models as embodied brain},
  author={Luo, Yulin and Fan, Chun-Kai and Dong, Menghang and Shi, Jiayu and Zhao, Mengdi and Zhang, Bo-Wen and Chi, Cheng and Liu, Jiaming and Dai, Gaole and Zhang, Rongyu and others},
  journal={arXiv preprint arXiv:2510.17801},
  year={2025}
}

@article{erqa,
  title={Gemini robotics: Bringing ai into the physical world},
  author={Team, Gemini Robotics and Abeyruwan, Saminda and Ainslie, Joshua and Alayrac, Jean-Baptiste and Arenas, Montserrat Gonzalez and Armstrong, Travis and Balakrishna, Ashwin and Baruch, Robert and Bauza, Maria and Blokzijl, Michiel and others},
  journal={arXiv preprint arXiv:2503.20020},
  year={2025}
}

@article{egoplan,
  title={Egoplan-bench: Benchmarking multimodal large language models for human-level planning},
  author={Chen, Yi and Ge, Yuying and Ge, Yixiao and Ding, Mingyu and Li, Bohao and Wang, Rui and Xu, Ruifeng and Shan, Ying and Liu, Xihui},
  journal={International Journal of Computer Vision},
  volume={134},
  number={3},
  pages={118},
  year={2026},
  publisher={Springer}
}

@inproceedings{multiplan+,
  title={EmbodiedAgent: A Scalable Hierarchical Approach to Overcome Practical Challenge in Multi-Robot Control},
  author={Wan, Hanwen and Chen, Yifei and Deng, Yixuan and Wei, Zeyu and Li, Dongrui and Lin, Zexin and Wu, Donghao and Cheng, Jiu and Ji, Xiaoqiang},
  booktitle={2025 IEEE/RSJ International Conference on Intelligent Robots and Systems (IROS)},
  pages={12140--12146},
  year={2025},
  organization={IEEE}
}

@inproceedings{roboafford,
  title={Roboafford: A dataset and benchmark for enhancing object and spatial affordance learning in robot manipulation},
  author={Tang, Yingbo and Zhang, Lingfeng and Zhang, Shuyi and Zhao, Yinuo and Hao, Xiaoshuai},
  booktitle={Proceedings of the 33rd ACM International Conference on Multimedia},
  pages={12706--12713},
  year={2025}
}

@inproceedings{pixmopoints,
  title={Molmo and pixmo: Open weights and open data for state-of-the-art vision-language models},
  author={Deitke, Matt and Clark, Christopher and Lee, Sangho and Tripathi, Rohun and Yang, Yue and Park, Jae Sung and Salehi, Mohammadreza and Muennighoff, Niklas and Lo, Kyle and Soldaini, Luca and others},
  booktitle={Proceedings of the Computer Vision and Pattern Recognition Conference},
  pages={91--104},
  year={2025}
}

@inproceedings{refcoco,
  title={Modeling context in referring expressions},
  author={Yu, Licheng and Poirson, Patrick and Yang, Shan and Berg, Alexander C and Berg, Tamara L},
  booktitle={European conference on computer vision},
  pages={69--85},
  year={2016},
  organization={Springer}
}

@inproceedings{rt2,
  title={Rt-2: Vision-language-action models transfer web knowledge to robotic control},
  author={Zitkovich, Brianna and Yu, Tianhe and Xu, Sichun and Xu, Peng and Xiao, Ted and Xia, Fei and Wu, Jialin and Wohlhart, Paul and Welker, Stefan and Wahid, Ayzaan and others},
  booktitle={Conference on Robot Learning},
  pages={2165--2183},
  year={2023},
  organization={PMLR}
}

@misc{mimo,
      title={MiMo-Embodied: X-Embodied Foundation Model Technical Report}, 
      author={Xiaomi Embodied Intelligence Team},
      year={2025},
      eprint={2511.16518},
      archivePrefix={arXiv},
      primaryClass={cs.RO},
      url={https://arxiv.org/abs/2511.16518}, 
}

@article{hy,
title={HY-Embodied-0.5: Embodied Foundation Models for Real-World Agents},
author={Tencent Robotics X and HY Vision Team},
journal={arXiv preprint arXiv:2604.07430},
year={2026}
}

@article{internvla_a1,
  title={InternVLA-A1: Unifying Understanding, Generation and Action for Robotic Manipulation},
  author={Cai, Junhao and Cai, Zetao and Cao, Jiafei and Chen, Yilun and He, Zeyu and Jiang, Lei and Li, Hang and Li, Hengjie and Li, Yang and Liu, Yufei and others},
  journal={arXiv preprint arXiv:2601.02456},
  year={2026}
}

@misc{lin2026,
      title={A Systematic Study of Data Modalities and Strategies for Co-training Large Behavior Models for Robot Manipulation}, 
      author={Fanqi Lin and Kushal Arora and Jean Mercat and Haruki Nishimura and Paarth Shah and Chen Xu and Mengchao Zhang and Mark Zolotas and Maya Angeles and Owen Pfannenstiehl and Andrew Beaulieu and Jose Barreiros},
      year={2026},
      eprint={2602.01067},
      archivePrefix={arXiv},
      primaryClass={cs.RO},
      url={https://arxiv.org/abs/2602.01067}, 
}

@article{vlm4vla,
  title={VLM4VLA: Revisiting Vision-Language-Models in Vision-Language-Action Models},
  author={Zhang, Jianke and Chen, Xiaoyu and Wang, Qiuyue and Li, Mingsheng and Guo, Yanjiang and Hu, Yucheng and Zhang, Jiajun and Bai, Shuai and Lin, Junyang and Chen, Jianyu},
  journal={arXiv preprint arXiv:2601.03309},
  year={2026}
}

@misc{interna1,
  title={InternData-A1},
  author={InternData-A1 contributors},
  howpublished={\url{https://github.com/InternRobotics/InternManip}},  year={2025}
}

@article{openvla,
    title={OpenVLA: An Open-Source Vision-Language-Action Model},
    author={{Moo Jin} Kim and Karl Pertsch and Siddharth Karamcheti and Ted Xiao and Ashwin Balakrishna and Suraj Nair and Rafael Rafailov and Ethan Foster and Grace Lam and Pannag Sanketi and Quan Vuong and Thomas Kollar and Benjamin Burchfiel and Russ Tedrake and Dorsa Sadigh and Sergey Levine and Percy Liang and Chelsea Finn},
    journal = {arXiv preprint arXiv:2406.09246},
    year={2024}
}

@misc{agibot,
  title        = {Introducing AgiBot World Colosseo: A Large-scale Manipulation Platform for Scalable and Intelligent Embodied Systems},
  author       = {Shi, Modi and Lu, Yuxiang and Wang, Huijie and Xie, Chengen and Bu, Qingwen},
  year         = {2025},
  month        = {March},
  howpublished = {\url{https://opendrivelab.com/AgiBot-World/}},
  note         = {Blog post},
}

@misc{cambrian,
      title={Cambrian-1: A Fully Open, Vision-Centric Exploration of Multimodal LLMs}, 
      author={Shengbang Tong and Ellis Brown and Penghao Wu and Sanghyun Woo and Manoj Middepogu and Sai Charitha Akula and Jihan Yang and Shusheng Yang and Adithya Iyer and Xichen Pan and Ziteng Wang and Rob Fergus and Yann LeCun and Saining Xie},
      year={2024},
      eprint={2406.16860},
      archivePrefix={arXiv},
      primaryClass={cs.CV},
      url={https://arxiv.org/abs/2406.16860}, 
}

@article{Qwen3-VL,
      title={Qwen3-VL Technical Report}, 
      author={Shuai Bai and Yuxuan Cai and Ruizhe Chen and Keqin Chen and Xionghui Chen and Zesen Cheng and Lianghao Deng and Wei Ding and Chang Gao and Chunjiang Ge and Wenbin Ge and Zhifang Guo and Qidong Huang and Jie Huang and Fei Huang and Binyuan Hui and Shutong Jiang and Zhaohai Li and Mingsheng Li and Mei Li and Kaixin Li and Zicheng Lin and Junyang Lin and Xuejing Liu and Jiawei Liu and Chenglong Liu and Yang Liu and Dayiheng Liu and Shixuan Liu and Dunjie Lu and Ruilin Luo and Chenxu Lv and Rui Men and Lingchen Meng and Xuancheng Ren and Xingzhang Ren and Sibo Song and Yuchong Sun and Jun Tang and Jianhong Tu and Jianqiang Wan and Peng Wang and Pengfei Wang and Qiuyue Wang and Yuxuan Wang and Tianbao Xie and Yiheng Xu and Haiyang Xu and Jin Xu and Zhibo Yang and Mingkun Yang and Jianxin Yang and An Yang and Bowen Yu and Fei Zhang and Hang Zhang and Xi Zhang and Bo Zheng and Humen Zhong and Jingren Zhou and Fan Zhou and Jing Zhou and Yuanzhi Zhu and Ke Zhu},
	  journal={arXiv preprint arXiv:2511.21631},
      year={2025}
}

@article{octo,
  title={Octo: An open-source generalist robot policy},
  author={Team, Octo Model and Ghosh, Dibya and Walke, Homer and Pertsch, Karl and Black, Kevin and Mees, Oier and Dasari, Sudeep and Hejna, Joey and Kreiman, Tobias and Xu, Charles and others},
  journal={arXiv preprint arXiv:2405.12213},
  year={2024}
}

@article{pi0,
  title={$\pi_0$: A Vision-Language-Action Flow Model for General Robot Control},
  author={Black, Kevin and Brown, Noah and Driess, Danny and Esmail, Adnan and Equi, Michael and Finn, Chelsea and Fusai, Niccolo and Groom, Lachy and Hausman, Karol and Ichter, Brian and others},
  journal={arXiv preprint arXiv:2410.24164},
  year={2024}
}

@article{pi05,
  title={$\pi_{0.5}$: a Vision-Language-Action Model with Open-World Generalization},
  author={Intelligence, Physical and Black, Kevin and Brown, Noah and Darpinian, James and Dhabalia, Karan and Driess, Danny and Esmail, Adnan and Equi, Michael and Finn, Chelsea and Fusai, Niccolo and others},
  journal={arXiv preprint arXiv:2504.16054},
  year={2025}
}

@article{fast,
  title={Fast: Efficient action tokenization for vision-language-action models},
  author={Pertsch, Karl and Stachowicz, Kyle and Ichter, Brian and Driess, Danny and Nair, Suraj and Vuong, Quan and Mees, Oier and Finn, Chelsea and Levine, Sergey},
  journal={arXiv preprint arXiv:2501.09747},
  year={2025}
}

@article{egodex,
  title={EgoDex: Learning Dexterous Manipulation from Large-Scale Egocentric Video},
  author={Ryan Hoque and Peide Huang and David J. Yoon and Mouli Sivapurapu and Jian Zhang},
  journal={ArXiv},
  year={2025},
  volume={abs/2505.11709},
  url={https://api.semanticscholar.org/CorpusID:278739529}
}

@article{rt1,
  title={Rt-1: Robotics transformer for real-world control at scale},
  author={Brohan, Anthony and Brown, Noah and Carbajal, Justice and Chebotar, Yevgen and Dabis, Joseph and Finn, Chelsea and Gopalakrishnan, Keerthana and Hausman, Karol and Herzog, Alex and Hsu, Jasmine and others},
  journal={arXiv preprint arXiv:2212.06817},
  year={2022}
}

@article{fastwam,
  title={Fast-WAM: Do World Action Models Need Test-time Future Imagination?},
  author={Yuan, Tianyuan and Dong, Zibin and Liu, Yicheng and Zhao, Hang},
  journal={arXiv preprint arXiv:2603.16666},
  year={2026}
}

@article{joyra01,
  title={JoyAI-RA 0.1: A Foundation Model for Robotic Autonomy},
  author={Zhang, Tianle and Yuan, Zhihao and Chi, Dafeng and Liu, Peidong and Li, Dongwei and Hu, Kejun and Zhang, Likui and Nie, Junnan and Wei, Ziming and Chen, Zengjue and others},
  journal={arXiv preprint arXiv:2604.20100},
  year={2026}
}

@inproceedings{ego4d,
  title={Ego4d: Around the world in 3,000 hours of egocentric video},
  author={Grauman, Kristen and Westbury, Andrew and Byrne, Eugene and Chavis, Zachary and Furnari, Antonino and Girdhar, Rohit and Hamburger, Jackson and Jiang, Hao and Liu, Miao and Liu, Xingyu and others},
  booktitle={Proceedings of the IEEE/CVF conference on computer vision and pattern recognition},
  pages={18995--19012},
  year={2022}
}

@article{ACT,
  title={Learning fine-grained bimanual manipulation with low-cost hardware},
  author={Zhao, Tony Z and Kumar, Vikash and Levine, Sergey and Finn, Chelsea},
  journal={arXiv preprint arXiv:2304.13705},
  year={2023}
}

@article{gal,
  title={Galaxea open-world dataset and g0 dual-system vla model},
  author={Jiang, Tao and Yuan, Tianyuan and Liu, Yicheng and Lu, Chenhao and Cui, Jianning and Liu, Xiao and Cheng, Shuiqi and Gao, Jiyang and Xu, Huazhe and Zhao, Hang},
  journal={arXiv preprint arXiv:2509.00576},
  year={2025}
}

@article{egoscale,
  title={Egoscale: Scaling dexterous manipulation with diverse egocentric human data},
  author={Zheng, Ruijie and Niu, Dantong and Xie, Yuqi and Wang, Jing and Xu, Mengda and Jiang, Yunfan and Casta{\~n}eda, Fernando and Hu, Fengyuan and Tan, You Liang and Fu, Letian and others},
  journal={arXiv preprint arXiv:2602.16710},
  year={2026}
}

@article{ki,
  title={Knowledge insulating vision-language-action models: Train fast, run fast, generalize better},
  author={Driess, Danny and Springenberg, Jost and Ichter, Brian and Yu, Lili and Li-Bell, Adrian and Pertsch, Karl and Ren, Allen and Walke, Homer and Vuong, Quan and Shi, Lucy Xiaoyang and others},
  journal={Advances in Neural Information Processing Systems},
  volume={38},
  pages={102867--102888},
  year={2026}
}

@article{liu2025unified,
  title={Unified Embodied VLM Reasoning with Robotic Action via Autoregressive Discretized Pre-training},
  author={Liu, Yi and Wang, Sukai and Wei, Dafeng and Cai, Xiaowei and Zhong, Linqing and Yang, Jiange and Ren, Guanghui and Zhang, Jinyu and Yao, Maoqing and Li, Chuankang and others},
  journal={arXiv preprint arXiv:2512.24125},
  year={2025}
}

@inproceedings{embodiedqa,
  title={Embodied question answering},
  author={Das, Abhishek and Datta, Samyak and Gkioxari, Georgia and Lee, Stefan and Parikh, Devi and Batra, Dhruv},
  booktitle={Proceedings of the IEEE conference on computer vision and pattern recognition},
  pages={1--10},
  year={2018}
}

@inproceedings{openeqa,
  title={Openeqa: Embodied question answering in the era of foundation models},
  author={Majumdar, Arjun and Ajay, Anurag and Zhang, Xiaohan and Putta, Pranav and Yenamandra, Sriram and Henaff, Mikael and Silwal, Sneha and Mcvay, Paul and Maksymets, Oleksandr and Arnaud, Sergio and others},
  booktitle={Proceedings of the IEEE/CVF conference on computer vision and pattern recognition},
  pages={16488--16498},
  year={2024}
}

@inproceedings{robovqa,
  title={Robovqa: Multimodal long-horizon reasoning for robotics},
  author={Sermanet, Pierre and Ding, Tianli and Zhao, Jeffrey and Xia, Fei and Dwibedi, Debidatta and Gopalakrishnan, Keerthana and Chan, Christine and Dulac-Arnold, Gabriel and Maddineni, Sharath and Joshi, Nikhil J and others},
  booktitle={2024 IEEE International Conference on Robotics and Automation (ICRA)},
  pages={645--652},
  year={2024},
  organization={IEEE}
}

@inproceedings{physbench,
  title={Physbench: Benchmarking and enhancing vision-language models for physical world understanding},
  author={Chow, Wei and Mao, Jiageng and Li, Boyi and Seita, Daniel and Campagnolo Guizilini, Vitor and Wang, Yue},
  booktitle={International Conference on Learning Representations},
  volume={2025},
  pages={97959--98108},
  year={2025}
}

@inproceedings{physvlm,
  title={Physvlm: Enabling visual language models to understand robotic physical reachability},
  author={Zhou, Weijie and Tao, Manli and Zhao, Chaoyang and Guo, Haiyun and Dong, Honghui and Tang, Ming and Wang, Jinqiao},
  booktitle={Proceedings of the Computer Vision and Pattern Recognition Conference},
  pages={6940--6949},
  year={2025}
}

@inproceedings{robobrain,
  title={Robobrain: A unified brain model for robotic manipulation from abstract to concrete},
  author={Ji, Yuheng and Tan, Huajie and Shi, Jiayu and Hao, Xiaoshuai and Zhang, Yuan and Zhang, Hengyuan and Wang, Pengwei and Zhao, Mengdi and Mu, Yao and An, Pengju and others},
  booktitle={Proceedings of the IEEE/CVF Conference on Computer Vision and Pattern Recognition},
  pages={1724--1734},
  year={2025}
}

@article{bert,
  title={BERT: a review of applications in natural language processing and understanding},
  author={Koroteev, Mikhail V},
  journal={arXiv preprint arXiv:2103.11943},
  year={2021}
}

@article{zhou,
  title={From perception to cognition: A survey of vision-language interactive reasoning in multimodal large language models},
  author={Zhou, Chenyue and Wang, Mingxuan and Ma, Yanbiao and Wu, Chenxu and Chen, Wanyi and Qian, Zhe and Liu, Xinyu and Zhang, Yiwei and Wang, Junhao and Xu, Hengbo and others},
  journal={arXiv preprint arXiv:2509.25373},
  year={2025}
}

@article{causalvqa,
  title={Causalvqa: A physically grounded causal reasoning benchmark for video models},
  author={Foss, Aaron and Evans, Chloe and Mitts, Sasha and Sinha, Koustuv and Rizvi, Ammar and Kao, Justine T},
  journal={arXiv preprint arXiv:2506.09943},
  year={2025}
}

@inproceedings{prismatic,
  title={Prismatic vlms: Investigating the design space of visually-conditioned language models},
  author={Karamcheti, Siddharth and Nair, Suraj and Balakrishna, Ashwin and Liang, Percy and Kollar, Thomas and Sadigh, Dorsa},
  booktitle={Forty-first International Conference on Machine Learning},
  year={2024}
}

@article{genrl,
  title={Genrl: Multimodal-foundation world models for generalization in embodied agents},
  author={Mazzaglia, Pietro and Verbelen, Tim and Dhoedt, Bart and Courville, Aaron and Rajeswar, Sai},
  journal={Advances in neural information processing systems},
  volume={37},
  pages={27529--27555},
  year={2024}
}

@article{phystoolbench,
  title={Phystoolbench: Benchmarking physical tool understanding for mllms},
  author={Zhang, Zixin and Chen, Kanghao and Lin, Xingwang and Jiang, Lutao and Zheng, Xu and Lyu, Yuanhuiyi and Guo, Litao and Li, Yinchuan and Chen, Ying-Cong},
  journal={arXiv preprint arXiv:2510.09507},
  year={2025}
}

@article{comprehensive,
  title={A comprehensive survey on visual question answering datasets and algorithms},
  author={Kabir, Raihan and Haque, Naznin and Islam, Md Saiful and others},
  journal={arXiv preprint arXiv:2411.11150},
  year={2024}
}

@article{being,
  title={Being-H0. 7: A Latent World-Action Model from Egocentric Videos},
  author={Luo, Hao and Zhang, Wanpeng and Feng, Yicheng and Zheng, Sipeng and Xu, Haiweng and Xu, Chaoyi and Xi, Ziheng and Fu, Yuhui and Lu, Zongqing},
  journal={arXiv preprint arXiv:2605.00078},
  year={2026}
}

@article{innon,
  title={In-N-On: Scaling Egocentric Manipulation with in-the-wild and on-task Data},
  author={Cai, Xiongyi and Qiu, Ri-Zhao and Chen, Geng and Wei, Lai and Liu, Isabella and Huang, Tianshu and Cheng, Xuxin and Wang, Xiaolong},
  journal={arXiv preprint arXiv:2511.15704},
  year={2025}
}


\clearpage
\appendix
\renewcommand{\thesection}{\Alph{section}}
\setcounter{section}{0}
\setcounter{figure}{0}
\setcounter{table}{0}
\renewcommand{\thefigure}{A\arabic{figure}}
\renewcommand{\thetable}{A\arabic{table}}

\begin{center}
{\Large\bfseries Appendix}\\[6pt]
\end{center}

\vspace{0.5em}
This appendix provides additional details and experimental results referenced in the main paper. ~\Cref{app:embodied_vqa} compares related embodied VQA datasets and benchmarks,~\cref{app:generator} documents the data generators used to construct CometData,~\cref{app:keyframe} describes the keyframe extraction algorithm,~\cref{app:real} contains further real‑robot experiment details, and~\cref{app:rt-pertask} reports per‑task RoboTwin success rates together with action analysis and per‑domain correlation results.


\section{Comparisons of Embodied VQA Datasets and Benchmarks}
\label{app:embodied_vqa}

\Cref{tab:data_source} and~\cref{tab:reasoning} provide a detailed comparison of representative embodied VQA datasets and benchmarks discussed in~\cref{sec:related_phyvqa}.~\Cref{tab:data_source} categorizes each dataset according to its data sources, distinguishing between robot-centric data (teleoperation, robot simulation, ego-centric robot views) and non-robot data (common web images, non-robot simulation).~\Cref{tab:reasoning} summarizes the reasoning capabilities covered by each benchmark, including spatial, planning, dynamics, error detection and recovery, affordance reasoning, and tool use, together with the scale of each benchmark where available.

\begin{table}[h]
\centering
\caption{Data source composition of representative embodied VQA datasets. For each dataset, we categorize its data sources into robot data (teleoperation, robot simulation, robot ego-centric view) and non-robot data (common web images, non-robot simulation).}
\label{tab:data_source}
\resizebox{\textwidth}{!}{%
\begin{tabular}{lcccccc}
\toprule
\multirow{2}{*}{Dataset} & \multirow{2}{*}{Visual cues} & \multicolumn{3}{c}{Robot Data} & \multicolumn{2}{c}{Non-Robot Data} \\
\cmidrule(lr){3-5} \cmidrule(lr){6-7}
& & Tele & Sim (Robot) & Ego (Robot) & Common & Sim (Non-Robot) \\
\midrule
RoboVQA~\cite{robovqa} & \full & \full & \none & \full & \none & \none \\
RefCOCO~\cite{refcoco} & \full & \none & \none & \none & \full & \none \\
RoboPoint~\cite{robopoint} & \full & \none & \none & \none & \none & \full \\
PixMo-Points~\cite{pixmopoints} & \none & \none & \none & \none & \full & \none \\
RefSpatial~\cite{refspatial} & \full & \none & \none & \none & \full & \full \\
RoboAfford~\cite{roboafford} & \full & \none & \full & \none & \full & \none \\
Robo2VLM~\cite{robo2vlm} & \none & \full & \none & \none & \none & \none \\
MultiPlan~\cite{multiplan, multiplan+} & \none & \none & \none & \none & \full & \none \\
Cosmos-Reason1~\cite{cosmosreason1} & \full & \full & \none & \full & \none & \parcirc \\
CometData (Ours) & \none & \full & \full & \full & \none & \none \\
\bottomrule
\addlinespace[3pt]
\multicolumn{7}{l}{\footnotesize \full~Fully Supported \quad \parcirc~Partially Supported \quad \none~Not Supported} \\
\end{tabular}}
\end{table}

\begin{table}[h]
\centering
\caption{Reasoning capability coverage of embodied VQA benchmarks. Scale indicates the number of question-answer pairs. Each benchmark is evaluated across six reasoning categories: Spatial, Planning, Dynamics, Error Detection \& Recovery, Affordance, and Tool Use.}
\label{tab:reasoning}
\begin{tabular}{lccccccc}
\toprule
\multirow{2}{*}{Benchmark} & \multirow{2}{*}{Scale} & \multicolumn{6}{c}{Reasoning Categories} \\
\cmidrule(lr){3-8}
& & Spatial & Plan & Dyn & Error & Afford & Tool \\
\midrule
RoboSpatial~\cite{robospatial} & 6,350 & \full & \none & \none & \none & \none & \none \\
EgoPlan~\cite{egoplan} & 1,584 & \none & \full & \none & \none & \none & \none \\
Cosmos-Reason1~\cite{cosmosreason1} & 1,205 & \full & \full & \none & \parcirc & \none & \none \\
ERQA~\cite{erqa} & 400 & \none & \full & \none & \parcirc & \none & \none \\
RoboVQA~\cite{robovqa} & 1,000 & \full & \full & \full & \none & \full & \none \\
RoboBench~\cite{robobench} & 6,092 & \full & \full & \none & \parcirc & \none & \none \\
ERIQ~\cite{liu2025unified} & 6,052 & \full & \full & \none & \full & \none & \none \\
CometBench (Ours) & -- & \full & \full & \full & \full & \full & \full \\
\bottomrule
\addlinespace[3pt]
\multicolumn{8}{l}{\footnotesize \full~Fully Supported \quad \parcirc~Partially Supported \quad \none~Not Supported} \\
\end{tabular}
\end{table}

\section{Data Generators}
\label{app:generator}

This appendix documents the procedural data generators used to produce visual question answering instances from raw robot manipulation trajectories. Each generator consumes a trajectory sample composed of synchronized multi-view video frames, a tabular telemetry timeline, and high-level task annotations. From these inputs, a generator selects evidence frames, instantiates a question template, and queries a vision language model server to obtain a grounded answer. The keyframe selection policy and the physics state extractor have been described in the main text, so the following subsections focus on the supervisory role of each generator and provide a representative question for every one.

\subsection{Spatial Reasoning}
\paragraph{Spatial 3D Generator}
This generator supervises three-dimensional spatial reasoning by asking the model to describe the relative arrangement of two task-relevant objects in terms of front, back, left, etc. It consumes a triple of evidence frames spanning the manipulation episode together with their physics state summaries. Example question:
\begin{quote}
\texttt{Where is the bottle relative to the cup?}
\end{quote}

\paragraph{Multi-View Matching Generator}
This generator targets viewpoint-invariant object grounding. At a single anchor moment, it presents synchronized frames from at least two camera views and asks the model to identify the same manipulated object across them, treating cross-view occlusion and perspective distortion as the central challenge. This generator is only activated when dataset has multiple views. Example question:
\begin{quote}
\texttt{Across the synchronized views shown, identify the object the robot is currently manipulating and explain how the views correspond.}
\end{quote}

\paragraph{Timeline Matching Generator}
This generator combines cross view and temporal correspondence. Two views are sampled at two distinct anchor moments, yielding four frames presented in a fixed view by time order. The model must first match the target object across views at each moment and then track it between the two moments, supervising joint viewpoint and temporal grounding. Example question:
\begin{quote}
\texttt{Across these four frames covering two views and two moments, track the bottle and describe how its state changes between the two moments.}
\end{quote}

\subsection{Task Understanding}

\paragraph{Subtask Decomposition Generator}
This generator supervises task planning at the subtask level. Given a single image and the high-level task description, it asks the model to predict what the next subtask should be, optionally conditioned on the current subtask. The prompt forbids the model from referencing hidden metadata, so the prediction must be grounded in observable robot configuration, object placement and tool state. Example question:

\begin{quote}
\texttt{The overall task is: pour water from a bottle into a cup and hand it to the user. The current subtask is: grasp the bottle. What should be the next subtask?}
\end{quote}

\paragraph{Qualitative Progress Generator}
This generator supervises qualitative progress estimation. The model is asked to describe, in natural language, where the robot currently stands within the overall task, for example whether execution is just starting, advancing, or nearly complete. The hidden completion ratio and the history of executed subtasks are exposed only as latent context, and the model must justify its qualitative answer purely from visual evidence. Example question:

\begin{quote}
\texttt{The overall task is: pour water from a bottle into a cup and hand it to the user. What is the current progress of the task?}
\end{quote}

\paragraph{Quantitative Progress Generator}
This generator supervises quantitative progress estimation by asking the model to justify a numerical completion percentage from the visible scene. The completion ratio is defined as the average of a structural component, computed as the fraction of planned subtasks already entered, and a temporal component, computed as the relative position of the current frame within the episode. The model is required to argue why the visible scene is consistent with this hidden ratio. Example question:

\begin{quote}
\texttt{The overall task is: pour water from a bottle into a cup and hand it to the user. The current subtask is: grasp the bottle. What is the progress percentile of this subtask?}
\end{quote}

\subsection{Physics \& Dynamics}

\paragraph{Physics Understanding Generator}
This generator supervises causal physical reasoning. The model is shown a physically salient frame and asked to explain how gravity, contact and energy will determine the immediate next outcome, for example whether the held object will stay stable, slip, or fall. Example question:
\begin{quote}
\texttt{If the robot releases the bottle now, what physical outcome is most likely?}
\end{quote}

\paragraph{Object Characteristic Generator}
This generator supervises dense object centric description. For each evidence frame, the model is asked to enumerate strongly task relevant objects and characterize them along a fixed schema covering visual evidence, color, material, shape and size, surface, rigidity, weight class, tool function, canonical and task specific use, role in the task pipeline, expected robot interaction and likely failure modes. Example question:
\begin{quote}
\texttt{Describe the physical and tool related properties of the clothes. Is it heavy?}
\end{quote}

\paragraph{Motion Understanding Generator}
This generator supervises action recognition from short, contact centered trajectory windows. The model is presented with a stack of consecutive frames around a physically salient keyframe and is asked to infer which manipulation action is being executed in the window, link motion and contact to causality, and analysis stability and risk. Example question:
\begin{quote}
\texttt{Looking at these images, what manipulation action is the robot executing, and what contact and stability cues support your answer?}
\end{quote}

\paragraph{Motion Imagination Generator}
This generator supervises short horizon action prediction. Given the current physical state, the model is asked to output a single immediate action that safely advances the high-level task, in a fixed seven-dimensional schema $[\,x, y, z, r_x, r_y, r_z, g\,]$ encoding body frame translation, rotation and a binary gripper command. The model is expected to respond with descriptive answers like "move along the x negative axis" or "close the gripper". Example question:
\begin{quote}
\texttt{The overall task is: pour water from a bottle into a cup and hand it to the user. The current subtask is: grasp the bottle. Considering the physics state, what should the robot do next?"}
\end{quote}

\subsection{Error \& Tool}
\paragraph{Execution Error Generator}
This generator supervises error detection. The ground truth task description is deliberately corrupted by substituting a verb or object token with a different one drawn from a curated pool, and the model is asked to judge whether the perturbed description correctly describes the current scene. The expected response is an explicit refutation, an analysis of the discrepancy with respect to gripper status and spatial relations, and a justified statement of the correct task. Example question:
\begin{quote}
\texttt{The current subtask is predicted as 'push the bottle'. Is this correct? If not, point out the error and provide the correct subtask.}
\end{quote}

\paragraph{Error Recovery Generator}
This generator supervises recovery planning under counterfactual failure. Reusing the same perturbation operator, it presents a hypothetical failed task together with the ground truth task and the physics state, and asks the model to design a minimal safe recovery procedure that returns the robot to the correct task while respecting torque limits and workspace constraints. Example question:
\begin{quote}
\texttt{Suppose the robot mistakenly attempted to push the bottle instead of grasping it. What is a minimal safe recovery plan to resume the correct task?}
\end{quote}

\paragraph{Constraint Detection Generator}
This generator supervises reasoning about the physical and geometric constraints that must be respected for safe execution. A keyword router classifies the constraint into one of three types: direction constraints for tasks involving containers or doors that must remain in a particular orientation, force constraints for fragile or deformable objects that require compliant contact, and space constraints for obstacle avoidance and reachability. The model is asked to explain the relevant constraints, why they matter, and how to satisfy them. Example question:
\begin{quote}
\texttt{For the task 'pour water from the bottle into the cup', what orientation constraints must the robot respect, and why?}
\end{quote}

\paragraph{Tool Usage Generator}
This generator supervises tool selection and identification. An upstream perception middle layer first issues a separate vision language query to detect whether a tool is present in the scene and, if so, returns its bounding box. The generator then asks the model to identify the appropriate tool, locate it when visible, and explain why it is suitable for the task, in either a single image or a multi image variant that exposes the workspace from several temporal vantage points. Example question:
\begin{quote}
\texttt{Which tool in the scene is most appropriate for the current task, where is it located, and why is it suitable?}
\end{quote}

\subsection{Affordance}

\paragraph{Affordance Reasoning Generator}
This generator supervises grasp and interaction affordance prediction, that is, where on a target object the robot should make contact in order to perform a given action. Frames around gripper transitions are preferred as evidence, since they typically capture the moment of contact establishment or release. The model is asked to point out the contact location and to explain the rationale. Example question:
\begin{quote}
\texttt{Where on the bottle should the robot grasp in order to lift it, and why?}
\end{quote}

\paragraph{Object Function Generator}
This generator supervises functional understanding of objects. The model is asked to describe the canonical function of a target object, the way it is used by the robot, and the purpose it serves in the depicted scene, using the image as primary support and the telemetry summary as secondary. Example question:
\begin{quote}
\texttt{What is the typical function of the cup in this scene, and what role does it play in the manipulation pipeline?}
\end{quote}

\section{Keyframe extraction}
\label{app:keyframe}

At frame \(t\), let \(s_t\) and \(a_t\) denote the synchronized proprioceptive state and commanded action. Let \(\mathcal{G}\) be the set of proprioceptive groups, and define the per-group tracking residual as \(\Delta_t^g \triangleq a_t^g - s_t^g\). Based on these quantities, we compute the instantaneous Hamiltonian \(H_t\), command effort \(\mathcal{E}_t\), and dimensionless command-to-response ratio \(\mathcal{R}_t\) as defined in~\cref{eq:her}
\begin{gather}
H_t = \tfrac{1}{2}\sum_{g \in \mathcal{G}} \|\Delta_t^g\|_2^2 + m g_0 z_t^{\text{ee}}, \qquad
\mathcal{E}_t = \sum_{g \in \mathcal{G}} \|\Delta_t^g\|_2, \qquad
\mathcal{R}_t = \frac{H_t - H_{t-1}}{\mathcal{E}_t + \varepsilon},
\label{eq:her}
\end{gather}
where \(m\) is the manipulator mass, \(g_0\) is gravitational acceleration, \(z_t^{\text{ee}}\) is the end-effector height (set to zero if pose is not in a world frame), and \(\varepsilon>0\) is a small constant for numerical stability.

Keyframes are identified using three complementary criteria in ~\cref{eq:ca}, ~\cref{eq:cb}, and ~\cref{eq:cc}. We use \(\mathbb{I}[\cdot]\) as the indicator function (1 if true, 0 otherwise), with thresholds \(\tau_H\), \(\tau_g\), \(\tau_{\mathcal{E}}\), and \(\tau_{\mathcal{R}}\). Hamiltonian jump \(\mathcal{C}_{\text{A}}\) detects abrupt energy redistribution during contact formation, impact, or release. Gripper step \(\mathcal{C}_{\text{B}}\) captures grasp/release boundaries through the normalized aperture \(g_t \in [0,1]\) and its first difference \(\Delta g_t \triangleq g_t-g_{t-1}\), where \(\mathrm{Sustain}_{K_{\text{sus}}}(t)\) indicates that \(|\Delta g_t|>\tau_g\) has held for the most recent \(K_{\text{sus}}\) frames. Power anomaly \(\mathcal{C}_{\text{C}}\) flags phases where command effort is large but Hamiltonian change or response ratio is disproportionately small; \(\alpha\) controls the attenuated energy threshold in this criterion.

\begin{gather}
\mathcal{C}_{\text{A}}(t) = \mathbb{I}\!\left[\,|H_t - H_{t-1}| > \tau_H\,\right],
\label{eq:ca}
\\
\mathcal{C}_{\text{B}}(t) =
\mathbb{I}\!\left[\,|\Delta g_t| > \tau_g \wedge \mathrm{Sustain}_{K_{\text{sus}}}(t)\,\right]
\;\vee\;
\mathbb{I}\!\left[\,(g_{t-1}-\tfrac{1}{2})(g_t-\tfrac{1}{2}) < 0\,\right],
\label{eq:cb}
\\
\mathcal{C}_{\text{C}}(t) =
\mathbb{I}\!\left[\,\mathcal{E}_t > \tau_{\mathcal{E}} \wedge \big(|H_t - H_{t-1}| < \alpha \tau_H \;\vee\; |\mathcal{R}_t| < \tau_{\mathcal{R}}\big)\,\right].
\label{eq:cc}
\end{gather}

The final saliency score aggregates these indicators to prioritize physically salient frames:
\begin{equation}
\mathcal{S}_t =
\frac{|H_t - H_{t-1}|}{\tau_H}
+ w_A \mathcal{C}_{\text{A}}(t)
+ w_B \mathcal{C}_{\text{B}}(t)
+ w_C \mathcal{C}_{\text{C}}(t),
\label{eq:saliency}
\end{equation}
with saliency mixing weights \((w_A,w_B,w_C)=(1.0,1.5,1.3)\) in our implementation. Keyframes are obtained by retaining the top-\(K\) frames per episode, where \(K\) is the per-episode keyframe budget.

Each selected keyframe is paired with a structured descriptor that includes the triggered criteria, normalized energy change \(|\Delta H_t|/\tau_H\), aperture change \(\Delta g_t\), command effort \(\mathcal{E}_t\), response ratio \(\mathcal{R}_t\), and an inferred task phase (grasp/release/manipulation). This descriptor serves as shared conditioning for all downstream category-specific QA generators. All quantities are computed solely from \((s_t, a_t, g_t)\), requiring no direct force or tactile sensing and ensuring uniform applicability across embodiments and data sources.

\section{Model Training configurations}

The training is generally divided into two stages: pre-training and post-training. For both stages, actions are denoted with end-effector position $ee_{l/r} = [x_{l/r}, y_{l/r}, z_{l/r}, rx_{l/r}, ry_{l/r}, rz_{l/r}, g_{l/r}]$, where $x, y, z$ denote spatial coordinates, $rx, ry, rz$ denote rotation angle, and $g$ denotes gripper open/close. This design generates a total of 14 dimensions for bi-manual control commands. We padded the action space to $32$ dimensions for future adaptation. We use an action chunk size of $30$ and perform inference with 4 flow-matching steps. In addition, CometVLA is trained with full parameters with no modules frozen.

In the pre-training stage, the model is trained on a mixture of vision-language and vision-language-action data. The VLM data includes general VQA and CometData with a per-device batch size of 4. The VLA data uses a mixture of AgiBot-World-Beta, InternData-A1, EgoLive, and EgoDex with a per-device batch size of 8. The training runs for 100,000 steps with a cosine learning rate scheduler. The base learning rate is set to 3e-5 and the warmup steps are 5000. The loss weights for VLA and VLM tasks are 1.0 and 0.1 respectively. The model is optimized with AdamW and gradient checkpointing and mixed precision training are enabled. The framework employs DiT-B as the action model and Qwen3-VL-4B-Instruct for the vision backbone.

In the post-training stage, the model is initialized from the pretrained checkpoint and fine-tuned on the RoboTwin or real-robot-specific dataset. The per-device batch size for VLA data is 16. The base learning rate remains 3e-5 with 5000 warmup steps and a cosine scheduler with minimum learning rate. The loss scale keeps VLA at 1.0 and VLM at 0.1. The model is trained in a fully fine-tuned manner without freezing any modules. The action model uses DiT-B with isolated embeddings and the same optimizer and training techniques as in the pre-training stage..

\section{Details of Real-Robot Experiments}
\label{app:real}

\subsection{Experiment Configurations}
The workstation runs the robot DDS middleware and directly sends end-effector pose commands in the camera coordinate system to the robot actuators. The computation-intensive policy inference runs on a remote H200 server cluster, which serves as the server. We establish WebSocket connections between the server and client through port forwarding, enabling low-latency command streaming. This hybrid architecture separates cloud-based inference from edge control, where the H200 cluster handles high-concurrency model inference and the 5090 workstation manages real-time robot control and state feedback.

\subsection{Visualization of the Real-Robot Experiments}

This appendix provides a per-task visual gallery of the five real-robot manipulation tasks. Each task is illustrated by six uniformly sampled keyframes arranged left-to-right to show the execution progression, followed by the corresponding language instruction.

\vspace{0.5em}
\noindent\textbf{Task 1 -- Lotion.} \textit{The right arm picks up the lotion from the desk and places the lotion into the long strip-shaped compartment at the back of the top layer of the storage box.}\\[2pt]
\setlength{\tabcolsep}{1pt}
\renewcommand{\arraystretch}{0.2}
\begin{tabular}{cccccc}
    \includegraphics[width=0.155\textwidth]{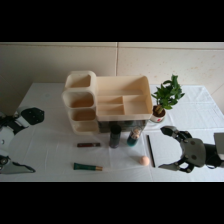} &
    \includegraphics[width=0.155\textwidth]{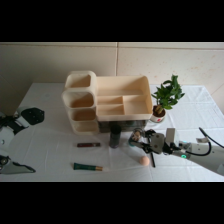} &
    \includegraphics[width=0.155\textwidth]{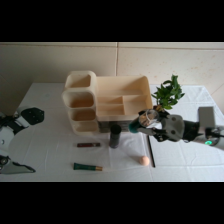} &
    \includegraphics[width=0.155\textwidth]{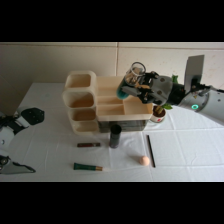} &
    \includegraphics[width=0.155\textwidth]{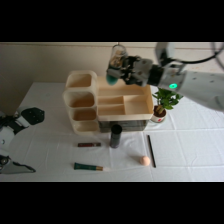} &
    \includegraphics[width=0.155\textwidth]{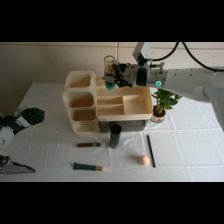} \\
\end{tabular}\\[4pt]

\vspace{0.8em}
\noindent\textbf{Task 2 -- Serum.} \textit{The right arm picks up the serum from the desk and places the serum into the long strip-shaped compartment at the back of the top layer of the storage box.}\\[2pt]
\setlength{\tabcolsep}{1pt}
\renewcommand{\arraystretch}{0.2}
\begin{tabular}{cccccc}
    \includegraphics[width=0.155\textwidth]{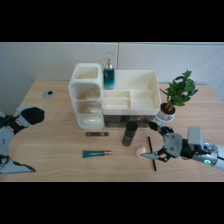} &
    \includegraphics[width=0.155\textwidth]{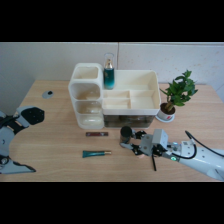} &
    \includegraphics[width=0.155\textwidth]{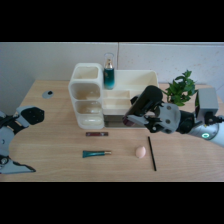} &
    \includegraphics[width=0.155\textwidth]{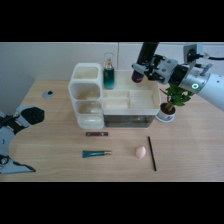} &
    \includegraphics[width=0.155\textwidth]{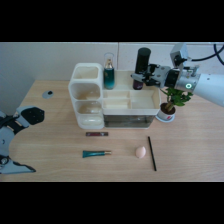} &
    \includegraphics[width=0.155\textwidth]{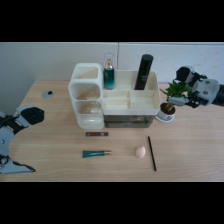} \\
\end{tabular}\\[4pt]

\vspace{0.8em}
\noindent\textbf{Task 3 -- Makeup Sponge.} \textit{The right arm picks up the makeup sponge from the desk and places the makeup sponge into the right compartment at the front of the top layer of the storage box.}\\[2pt]
\setlength{\tabcolsep}{1pt}
\renewcommand{\arraystretch}{0.2}
\begin{tabular}{cccccc}
    \includegraphics[width=0.155\textwidth]{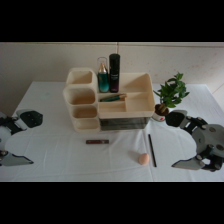} &
    \includegraphics[width=0.155\textwidth]{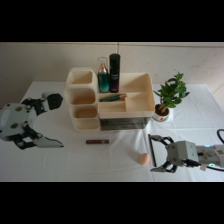} &
    \includegraphics[width=0.155\textwidth]{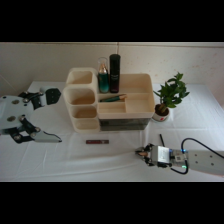} &
    \includegraphics[width=0.155\textwidth]{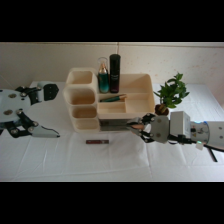} &
    \includegraphics[width=0.155\textwidth]{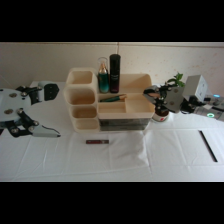} &
    \includegraphics[width=0.155\textwidth]{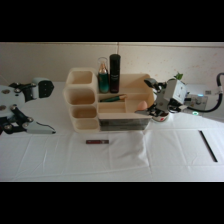} \\
\end{tabular}\\[4pt]

\vspace{0.8em}
\noindent\textbf{Task 4 -- BB Cream.} \textit{The left arm picks up the BB cream from the desk and places the BB cream into the middle compartment at the front of the top layer of the storage box.}\\[2pt]
\setlength{\tabcolsep}{1pt}
\renewcommand{\arraystretch}{0.2}
\begin{tabular}{cccccc}
    \includegraphics[width=0.155\textwidth]{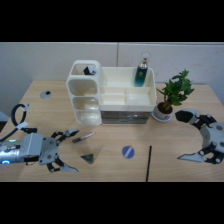} &
    \includegraphics[width=0.155\textwidth]{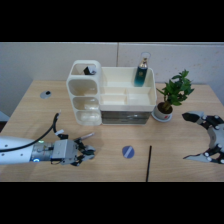} &
    \includegraphics[width=0.155\textwidth]{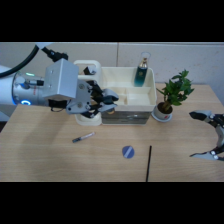} &
    \includegraphics[width=0.155\textwidth]{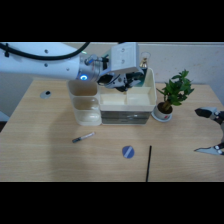} &
    \includegraphics[width=0.155\textwidth]{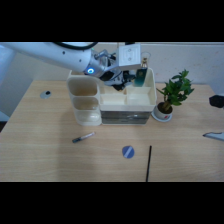} &
    \includegraphics[width=0.155\textwidth]{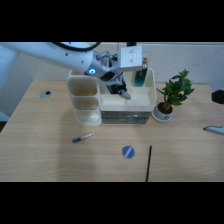} \\
\end{tabular}\\[4pt]

\vspace{0.8em}
\noindent\textbf{Task 5 -- Liquid Foundation.} \textit{The left arm picks up the liquid foundation from the desk and places the liquid foundation into the top open compartment on the left side of the storage box.}\\[2pt]
\setlength{\tabcolsep}{1pt}
\renewcommand{\arraystretch}{0.2}
\begin{tabular}{cccccc}
    \includegraphics[width=0.155\textwidth]{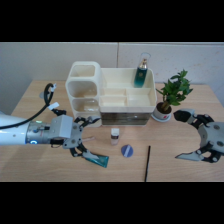} &
    \includegraphics[width=0.155\textwidth]{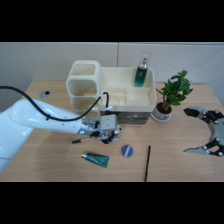} &
    \includegraphics[width=0.155\textwidth]{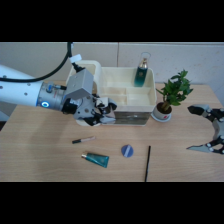} &
    \includegraphics[width=0.155\textwidth]{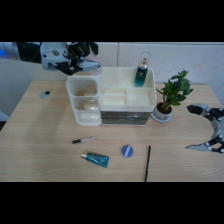} &
    \includegraphics[width=0.155\textwidth]{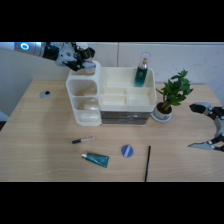} &
    \includegraphics[width=0.155\textwidth]{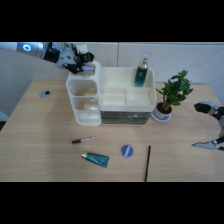} \\
\end{tabular}\\[4pt]

\subsection{Failure case analysis}

\begin{figure}[htbp]
    \centering
    \includegraphics[width=0.88\textwidth]{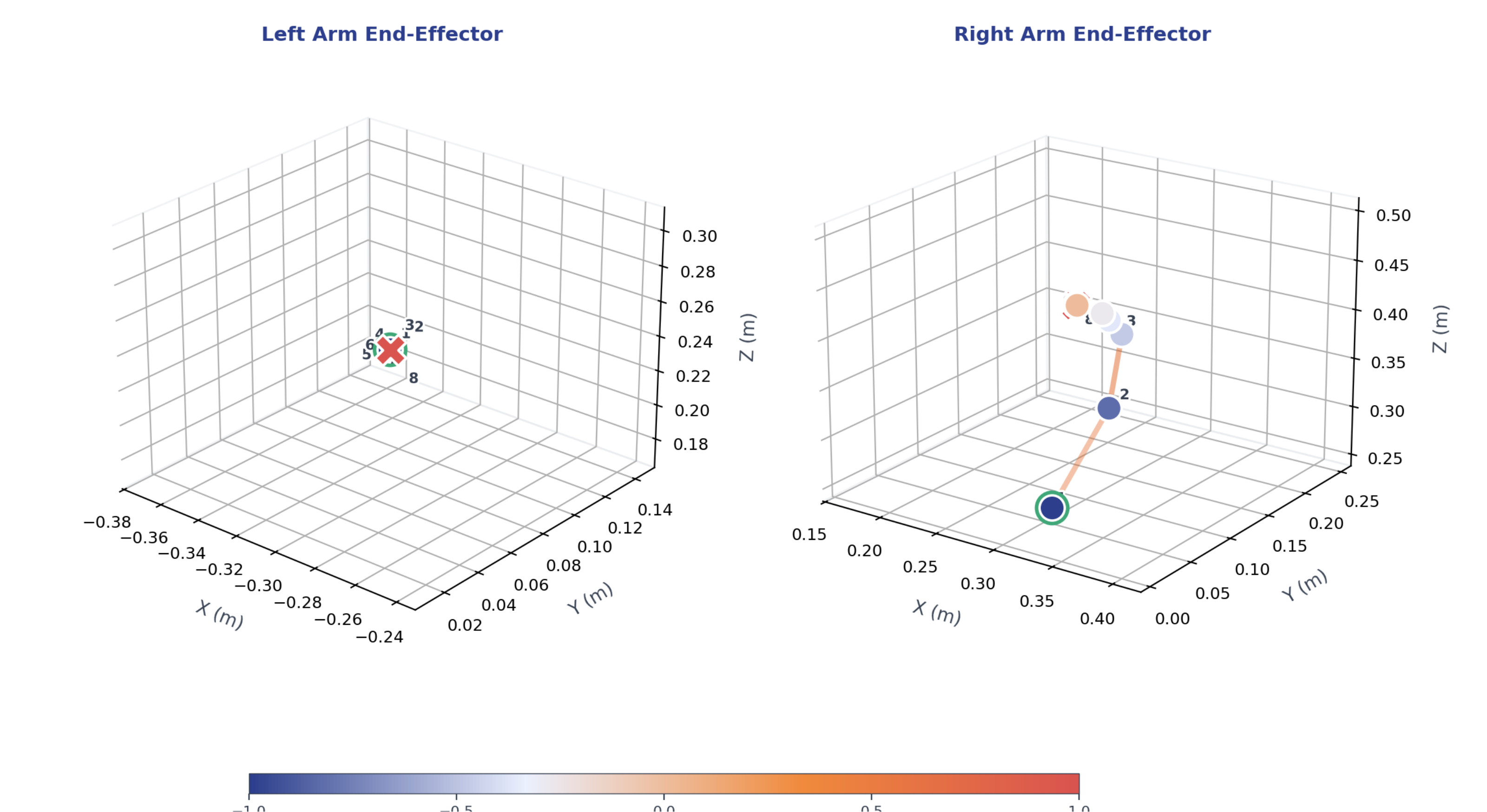}
    \caption{A failure case of lotion organization. }
    \label{fig:fail_real}
\end{figure}

We visualize a representative failure case from the real-robot experiments in~\cref{fig:fail_real}. The robot is instructed to pick up the lotion bottle with the right arm from the desk and place it into the long strip-shaped compartment located at the back of the top layer of the storage box. While the left arm correctly remains static as instructed, the right arm opens the gripper prematurely before reaching the lotion bottle, causing the object to be knocked over. The testers subsequently terminated the execution to avoid potential collisions and unsafe motions.

\section{Details of Evaluation on RoboTwin 2.0}

\subsection{Per-task Success Rate}
\label{app:rt-pertask}
We report the success rate of the full 50 tasks of CometVLA on RoboTwin 2.0 in~\cref{tab:robotwin}.

\begin{longtable}{lcc}
\caption{Per-task success counts}
\label{tab:robotwin} \\
\toprule
Task & Easy (Clean) & Hard (Randomized) \\
\midrule
\endfirsthead

\multicolumn{3}{c}{Continued: Per-task success counts} \\
\toprule
Task & Easy (Clean) & Hard (Randomized) \\
\midrule
\endhead

\bottomrule
\endfoot

adjust\_bottle & 100\% & 95\% \\
beat\_block\_hammer & 100\% & 96\% \\
blocks\_ranking\_rgb & 100\% & 97\% \\
blocks\_ranking\_size & 96\% & 79\% \\
click\_alarmclock & 71\% & 71\% \\
click\_bell & 94\% & 100\% \\
dump\_bin\_bigbin & 98\% & 80\% \\
grab\_roller & 99\% & 100\% \\
handover\_block & 83\% & 52\% \\
handover\_mic & 99\% & 97\% \\
hanging\_mug & 53\% & 54\% \\
lift\_pot & 100\% & 100\% \\
move\_can\_pot & 55\% & 97\% \\
move\_pillbottle\_pad & 90\% & 95\% \\
move\_playingcard\_away & 97\% & 92\% \\
move\_stapler\_pad & 91\% & 93\% \\
open\_laptop & 100\% & 98\% \\
open\_microwave & 92\% & 81\% \\
pick\_diverse\_bottles & 91\% & 65\% \\
pick\_dual\_bottles & 85\% & 72\% \\
place\_a2b\_left & 85\% & 97\% \\
place\_a2b\_right & 79\% & 86\% \\
place\_bread\_basket & 97\% & 98\% \\
place\_bread\_skillet & 100\% & 89\% \\
place\_burger\_fries & 96\% & 96\% \\
place\_can\_basket & 84\% & 51\% \\
place\_cans\_plasticbox & 99\% & 98\% \\
place\_container\_plate & 93\% & 100\% \\
place\_dual\_shoes & 94\% & 98\% \\
place\_empty\_cup & 99\% & 100\% \\
place\_fan & 82\% & 95\% \\
place\_mouse\_pad & 78\% & 82\% \\
place\_object\_basket & 85\% & 89\% \\
place\_object\_scale & 96\% & 83\% \\
place\_object\_stand & 94\% & 74\% \\
place\_phone\_stand & 86\% & 98\% \\
place\_shoe & 97\% & 84\% \\
press\_stapler & 95\% & 98\% \\
put\_bottles\_dustbin & 99\% & 91\% \\
put\_object\_cabinet & 71\% & 74\% \\
rotate\_qrcode & 83\% & 89\% \\
scan\_object & 80\% & 80\% \\
shake\_bottle & 92\% & 100\% \\
shake\_bottle\_horizontally & 98\% & 100\% \\
stack\_blocks\_three & 96\% & 97\% \\
stack\_blocks\_two & 97\% & 94\% \\
stack\_bowls\_three & 72\% & 86\% \\
stack\_bowls\_two & 83\% & 93\% \\
stamp\_seal & 87\% & 95\% \\
turn\_switch & 71\% & 90\% \\
Average & 89.24\% & 88.38\% \\

\end{longtable}

\subsection{Visualization of CometVLA on Simulation Tasks}
\label{app:rt-sim}
\begin{figure}[h]
    \centering
    \setlength{\tabcolsep}{1pt}
    \renewcommand{\arraystretch}{0.2}
    \begin{tabular}{ccccccc}
        \includegraphics[width=0.139\textwidth]{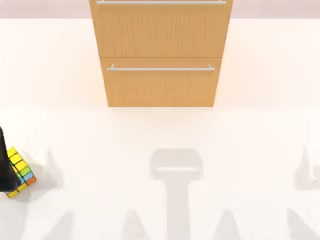} &
        \includegraphics[width=0.139\textwidth]{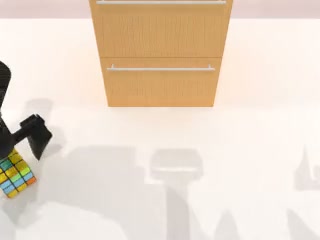} &
        \includegraphics[width=0.139\textwidth]{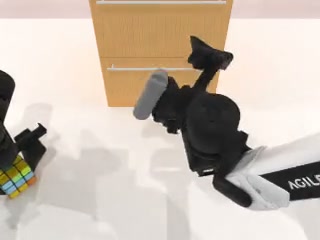} &
        \includegraphics[width=0.139\textwidth]{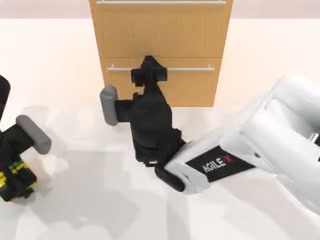} &
        \includegraphics[width=0.139\textwidth]{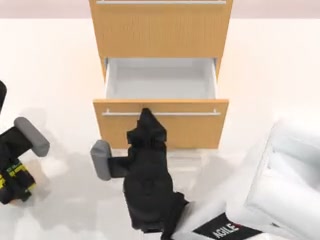} &
        \includegraphics[width=0.139\textwidth]{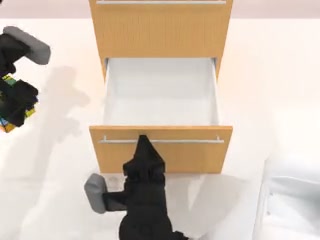} &
        \includegraphics[width=0.139\textwidth]{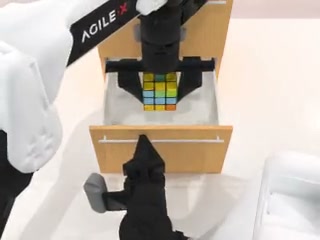} \\
        \includegraphics[width=0.139\textwidth]{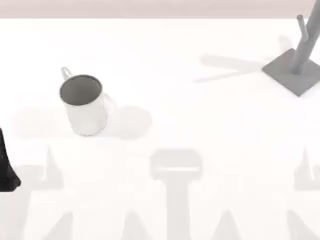} &
        \includegraphics[width=0.139\textwidth]{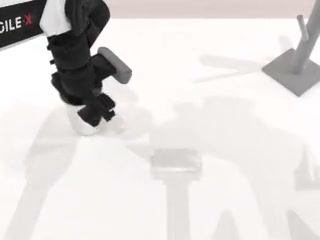} &
        \includegraphics[width=0.139\textwidth]{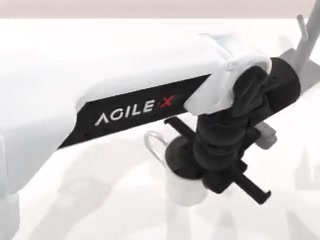} &
        \includegraphics[width=0.139\textwidth]{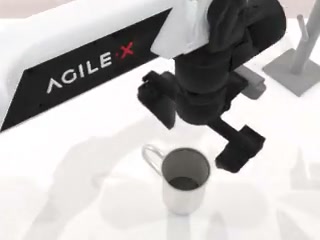} &
        \includegraphics[width=0.139\textwidth]{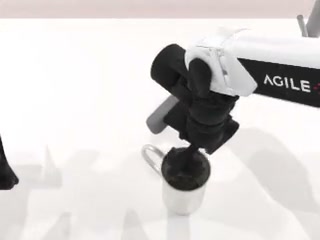} &
        \includegraphics[width=0.139\textwidth]{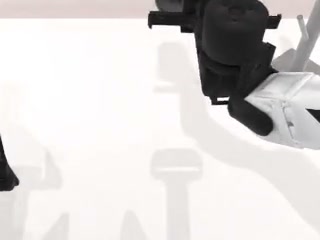} &
        \includegraphics[width=0.139\textwidth]{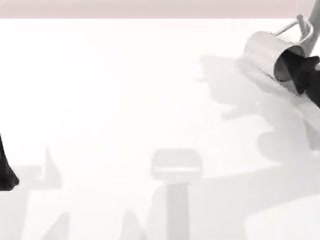} \\
        \includegraphics[width=0.139\textwidth]{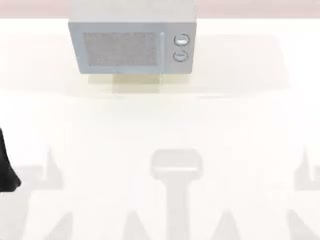} &
        \includegraphics[width=0.139\textwidth]{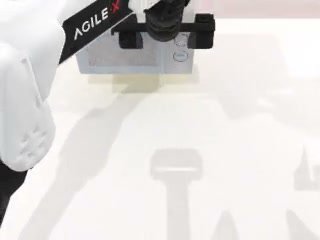} &
        \includegraphics[width=0.139\textwidth]{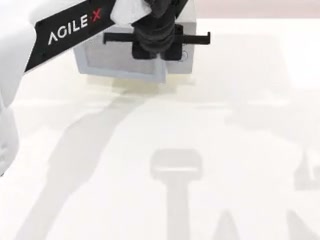} &
        \includegraphics[width=0.139\textwidth]{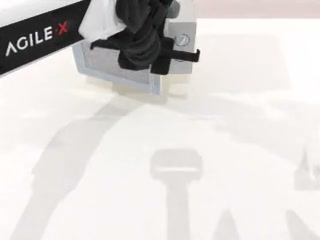} &
        \includegraphics[width=0.139\textwidth]{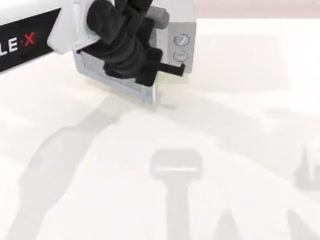} &
        \includegraphics[width=0.139\textwidth]{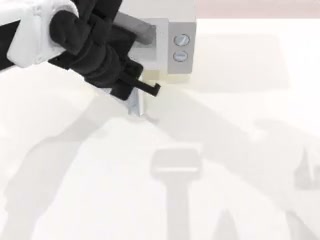} &
        \includegraphics[width=0.139\textwidth]{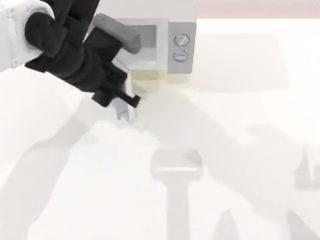} \\
        \includegraphics[width=0.139\textwidth]{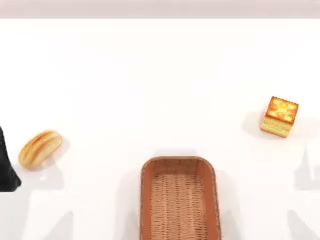} &
        \includegraphics[width=0.139\textwidth]{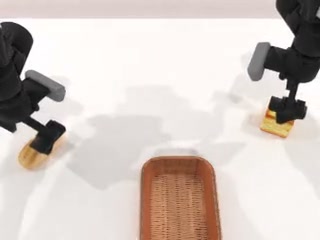} &
        \includegraphics[width=0.139\textwidth]{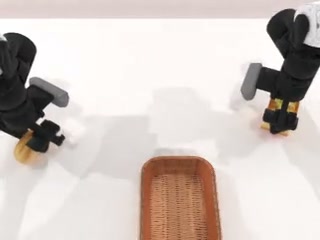} &
        \includegraphics[width=0.139\textwidth]{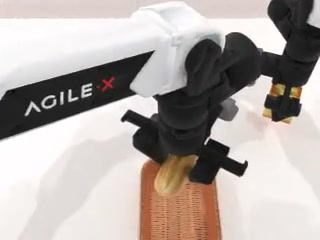} &
        \includegraphics[width=0.139\textwidth]{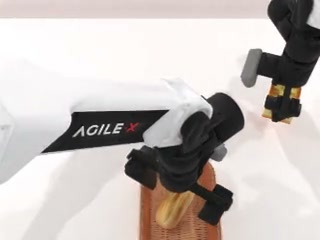} &
        \includegraphics[width=0.139\textwidth]{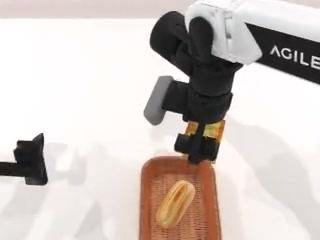} &
        \includegraphics[width=0.139\textwidth]{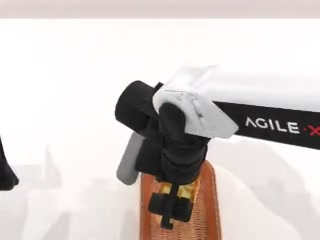} \\
        \includegraphics[width=0.139\textwidth]{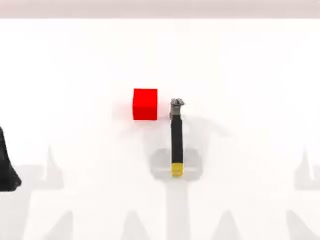} &
        \includegraphics[width=0.139\textwidth]{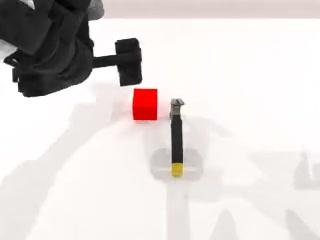} &
        \includegraphics[width=0.139\textwidth]{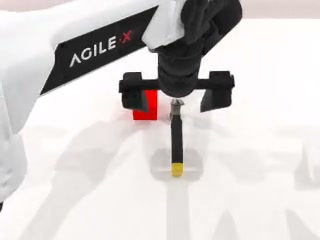} &
        \includegraphics[width=0.139\textwidth]{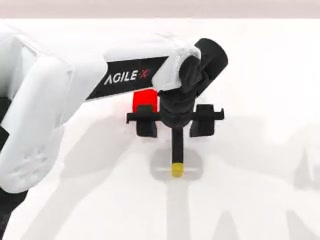} &
        \includegraphics[width=0.139\textwidth]{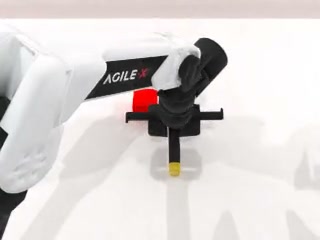} &
        \includegraphics[width=0.139\textwidth]{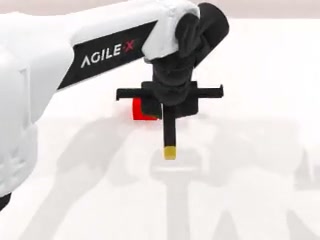} &
        \includegraphics[width=0.139\textwidth]{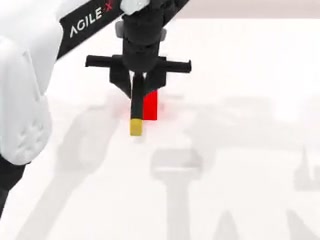} \\
        \includegraphics[width=0.139\textwidth]{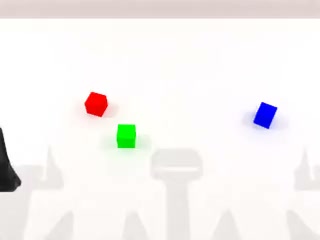} &
        \includegraphics[width=0.139\textwidth]{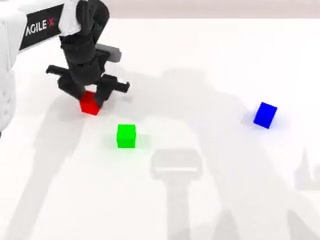} &
        \includegraphics[width=0.139\textwidth]{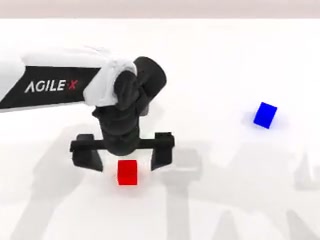} &
        \includegraphics[width=0.139\textwidth]{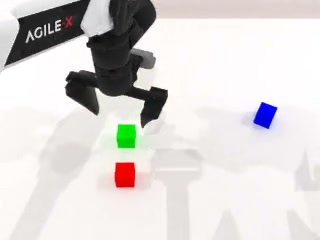} &
        \includegraphics[width=0.139\textwidth]{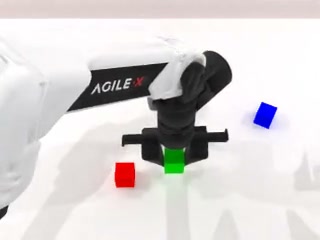} &
        \includegraphics[width=0.139\textwidth]{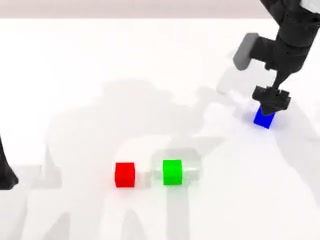} &
        \includegraphics[width=0.139\textwidth]{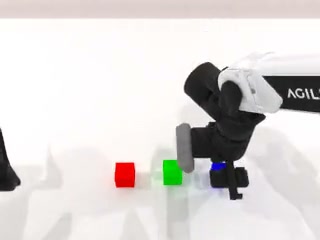} \\
        \includegraphics[width=0.139\textwidth]{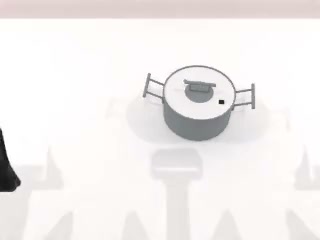} &
        \includegraphics[width=0.139\textwidth]{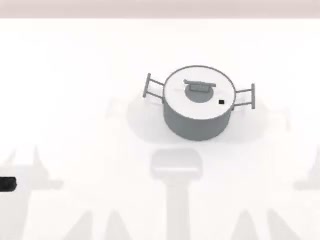} &
        \includegraphics[width=0.139\textwidth]{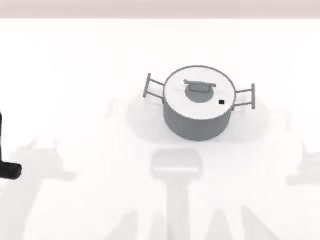} &
        \includegraphics[width=0.139\textwidth]{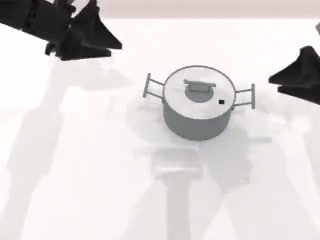} &
        \includegraphics[width=0.139\textwidth]{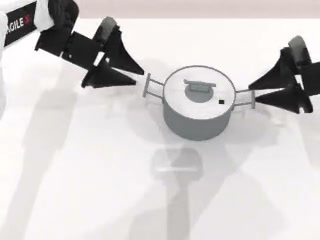} &
        \includegraphics[width=0.139\textwidth]{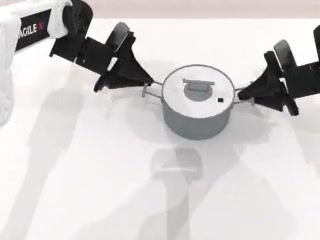} &
        \includegraphics[width=0.139\textwidth]{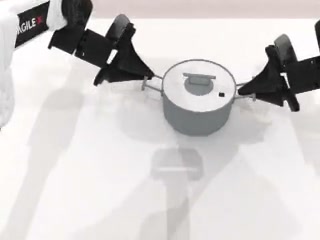} \\
        \includegraphics[width=0.139\textwidth]{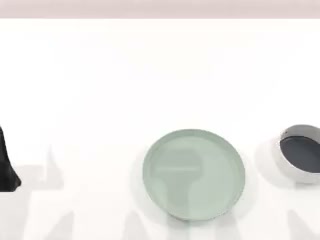} &
        \includegraphics[width=0.139\textwidth]{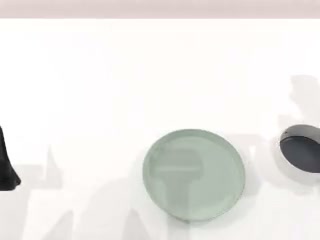} &
        \includegraphics[width=0.139\textwidth]{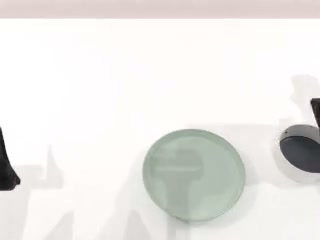} &
        \includegraphics[width=0.139\textwidth]{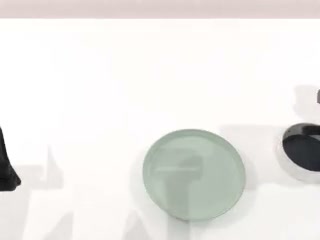} &
        \includegraphics[width=0.139\textwidth]{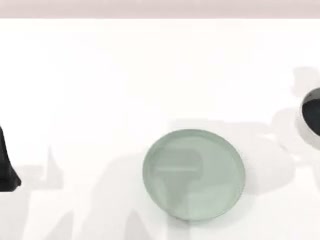} &
        \includegraphics[width=0.139\textwidth]{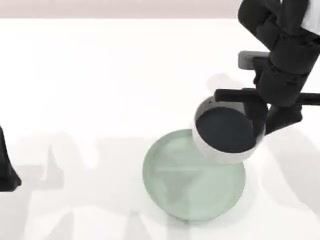} &
        \includegraphics[width=0.139\textwidth]{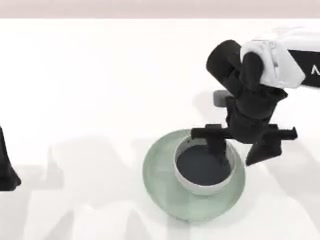} \\
        \includegraphics[width=0.139\textwidth]{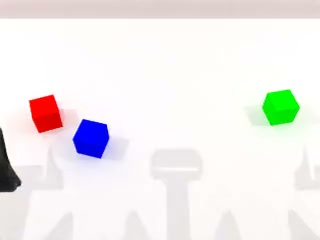} &
        \includegraphics[width=0.139\textwidth]{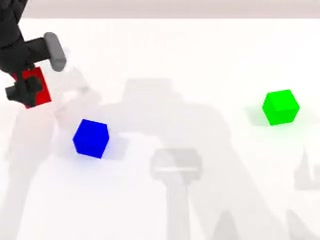} &
        \includegraphics[width=0.139\textwidth]{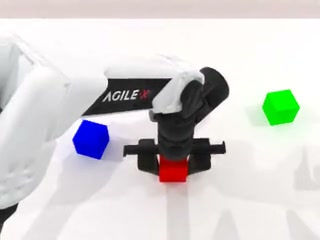} &
        \includegraphics[width=0.139\textwidth]{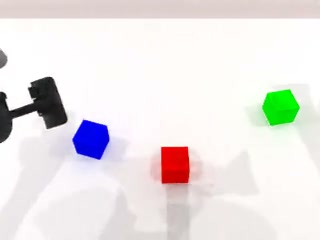} &
        \includegraphics[width=0.139\textwidth]{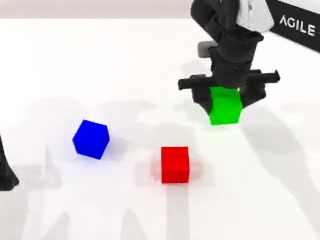} &
        \includegraphics[width=0.139\textwidth]{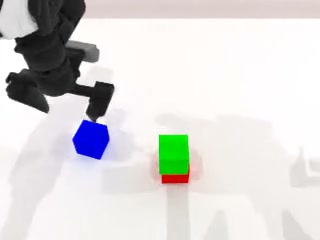} &
        \includegraphics[width=0.139\textwidth]{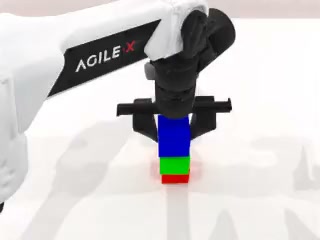} \\
        \includegraphics[width=0.139\textwidth]{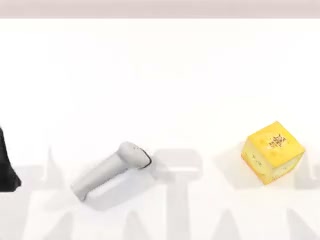} &
        \includegraphics[width=0.139\textwidth]{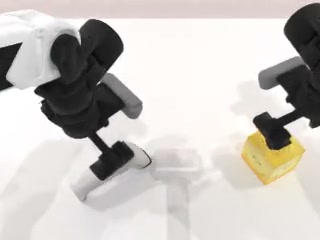} &
        \includegraphics[width=0.139\textwidth]{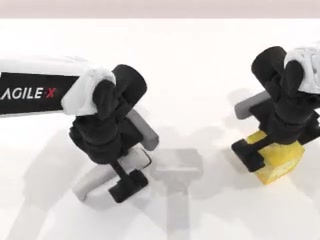} &
        \includegraphics[width=0.139\textwidth]{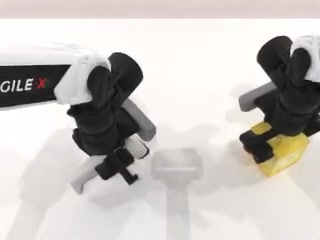} &
        \includegraphics[width=0.139\textwidth]{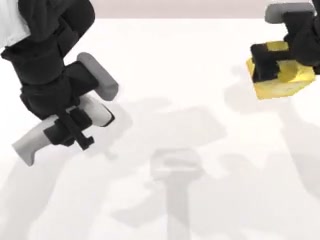} &
        \includegraphics[width=0.139\textwidth]{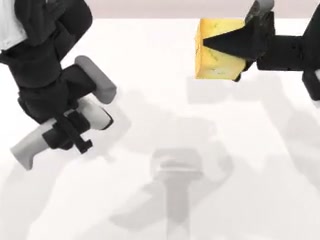} &
        \includegraphics[width=0.139\textwidth]{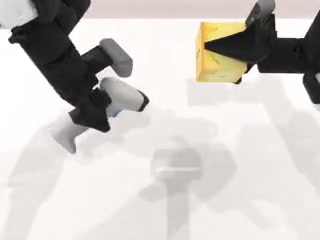} \\
    \end{tabular}
    \caption{Per-episode qualitative previews. Each row shows sampled frames from one rollout during the RoboTwin 2.0 Easy evaluation.}
    \label{fig:rollout_preview}
\end{figure}

We provide sampled rollout frames from RoboTwin 2.0 simulations in ~\cref{fig:rollout_preview}.

\subsection{Action Analysis}
\label{app:rt-action}
During the RoboTwin benchmark evaluation, we record the full 14-dimensional action sequence produced by our policy to diagnose the quality and temporal structure of generated robot motions. This appendix provides a complete visualization of a representative evaluation episode, including raw action time series, 3D end-effector trajectories, frequency-domain analysis, and gripper actuation events. These figures complement the quantitative success-rate metrics reported in the main text by revealing that our policy outputs smooth and physically plausible trajectories with minimal high-frequency jitter, while also demonstrating well-coordinated temporal alignment between the two grippers during bi-manual manipulation.

\begin{figure}[htbp]
    \centering
    \includegraphics[width=\textwidth]{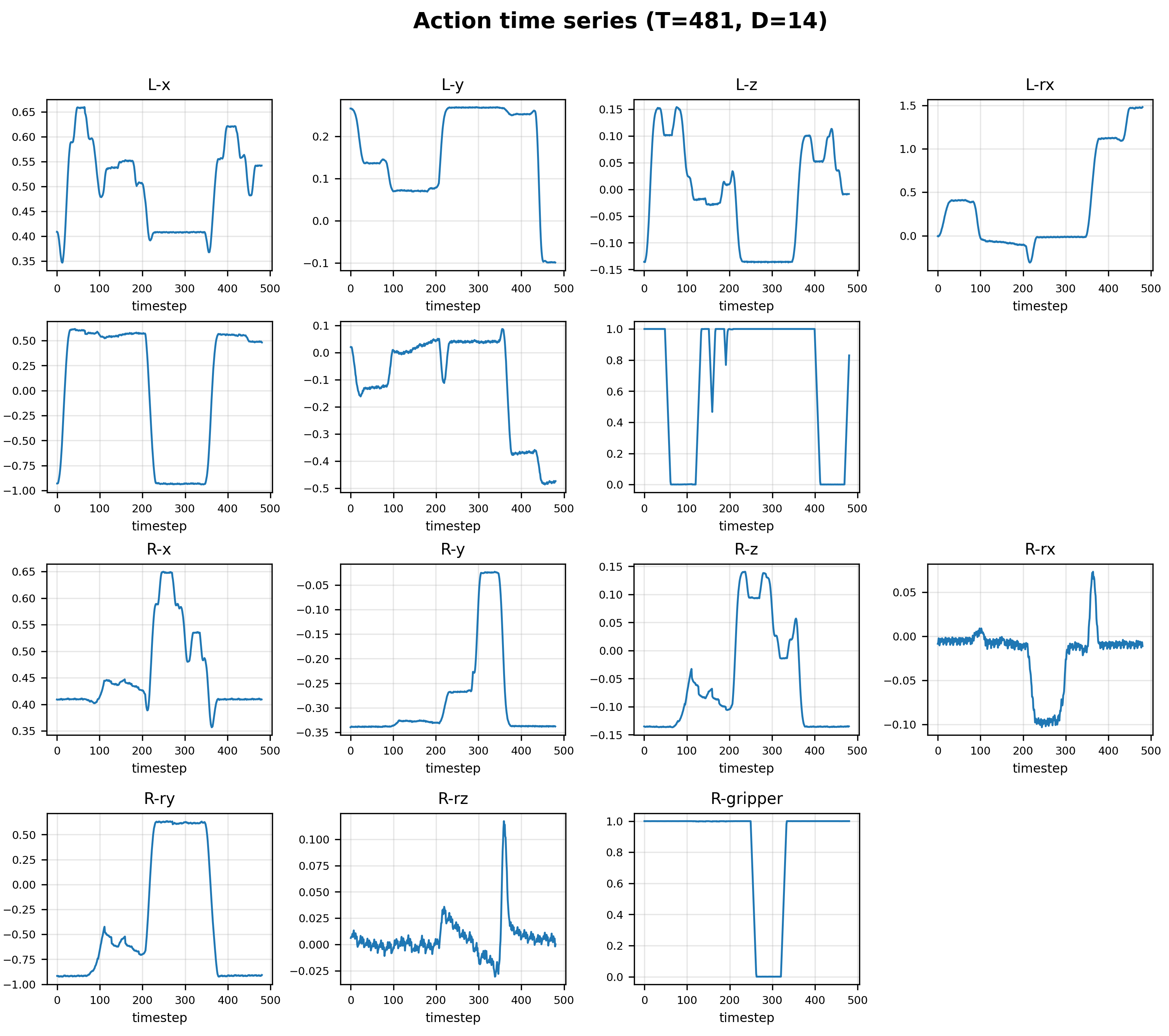}
    \caption{Complete action time series of a RoboTwin evaluation episode (timestep $T=481$, action dimension $D=14$). Step-like transitions and plateaus correspond to gripper actuation events and target-reaching phases.
    }
    \label{fig:app_action_timeseries}
\end{figure}

\begin{figure}[htbp]
    \centering
    \includegraphics[width=0.88\textwidth]{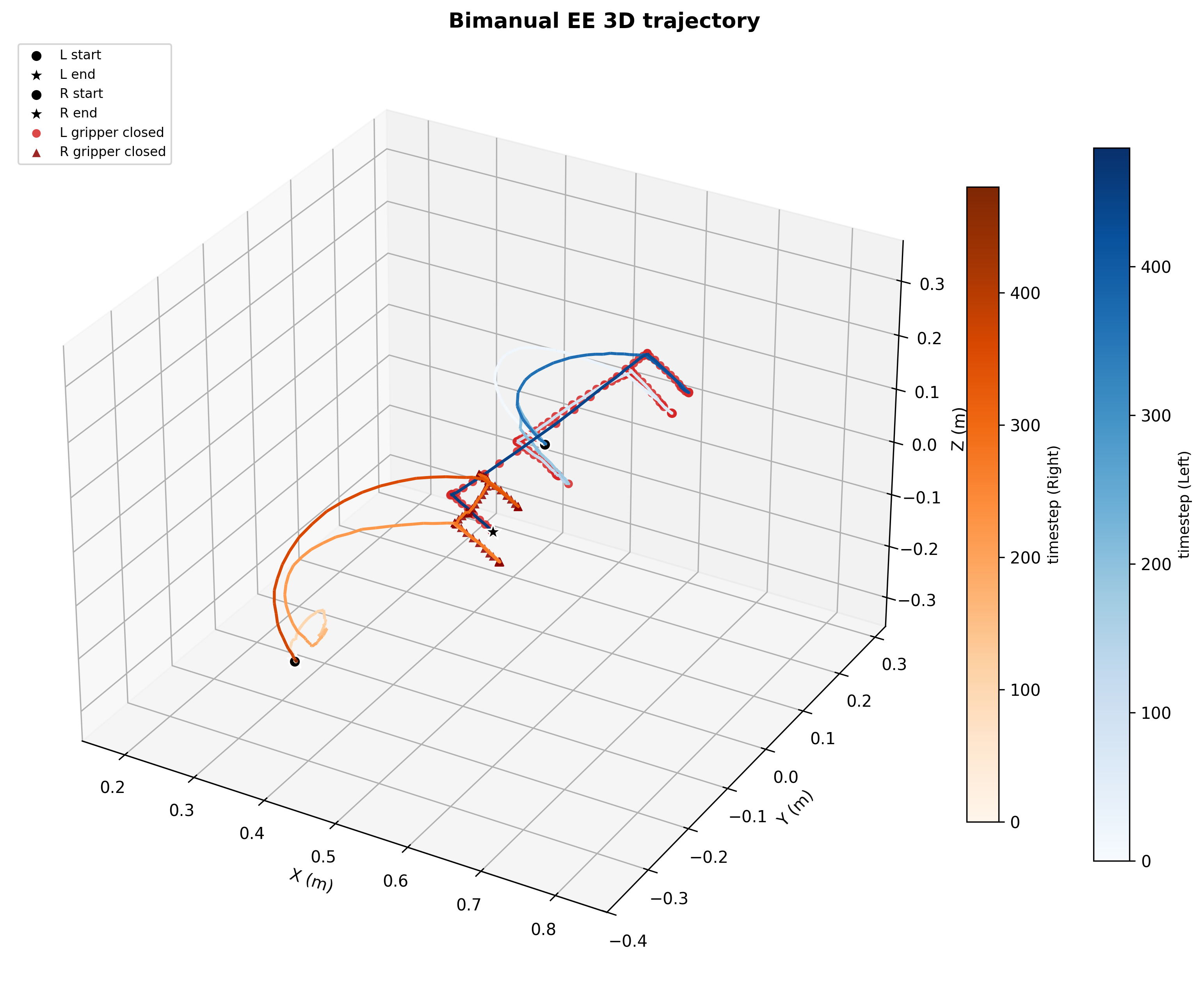}
    \caption{Three-dimensional end-effector trajectories of both arms. The blue and orange curves denote the left and right end-effector paths, respectively. Circles ($\bullet$) and stars ($\star$) mark the starting and ending poses. Red dots and triangles highlight timesteps where the left and right grippers are closed.
    }
    \label{fig:app_ee_trajectory}
\end{figure}

\begin{figure}[htbp]
    \centering
    \includegraphics[width=\textwidth]{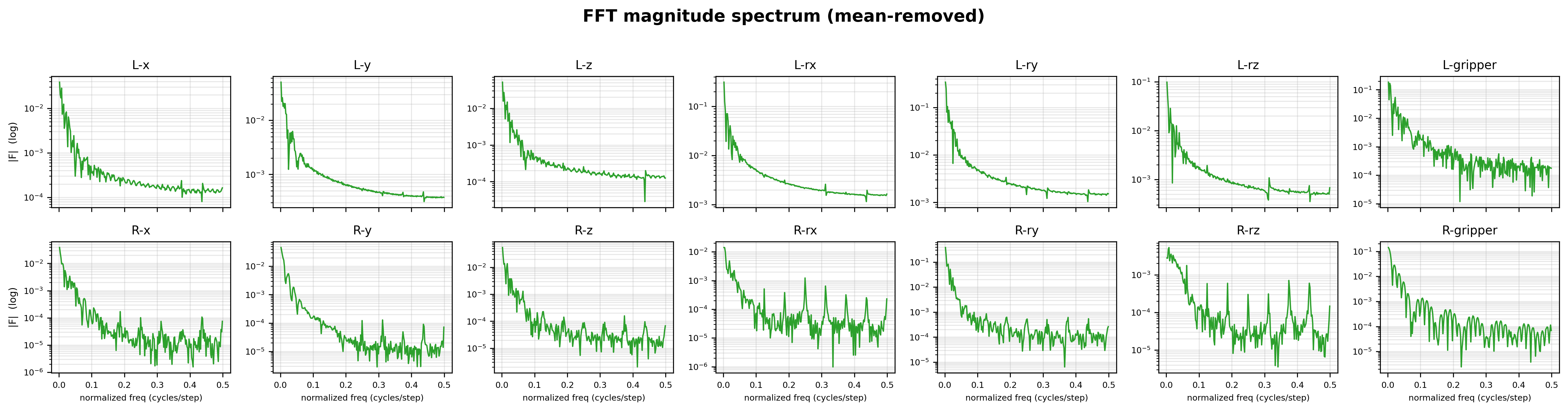}
    \caption{Mean-removed FFT magnitude spectrum for each of the 14 action dimensions. The vertical axis is logarithmic. Most signal energy is concentrated at low normalized frequencies ($<<0.1$ cycles/step), indicating that the manipulation motion is predominantly smooth and quasi-static,  with relatively little high-frequency jitter.
    }
    \label{fig:app_fft_spectrum}
\end{figure}

\begin{figure}[htbp]
    \centering
    \includegraphics[width=\textwidth]{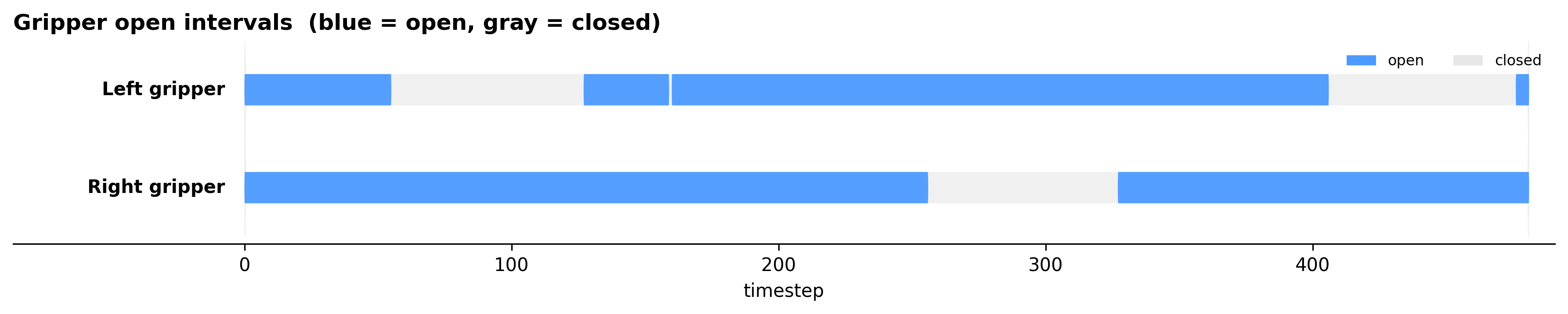}
    \caption{%
    Gripper open/close event timeline during the evaluation episode. Blue segments indicate open states and gray segments indicate closed states. The left and right grippers exhibit asynchronous actuation patterns, reflecting the coordinated bi-manual manipulation strategy required by the task. The short closed gap within the left gripper open interval is transient actuator jitter and recovers instantly.
}
    \label{fig:app_gripper_events}
\end{figure}

\subsection{Per-domain Correlation Analysis}
\label{app:percorr}
\begin{figure*}[t]
\centering
\includegraphics[width=\textwidth]{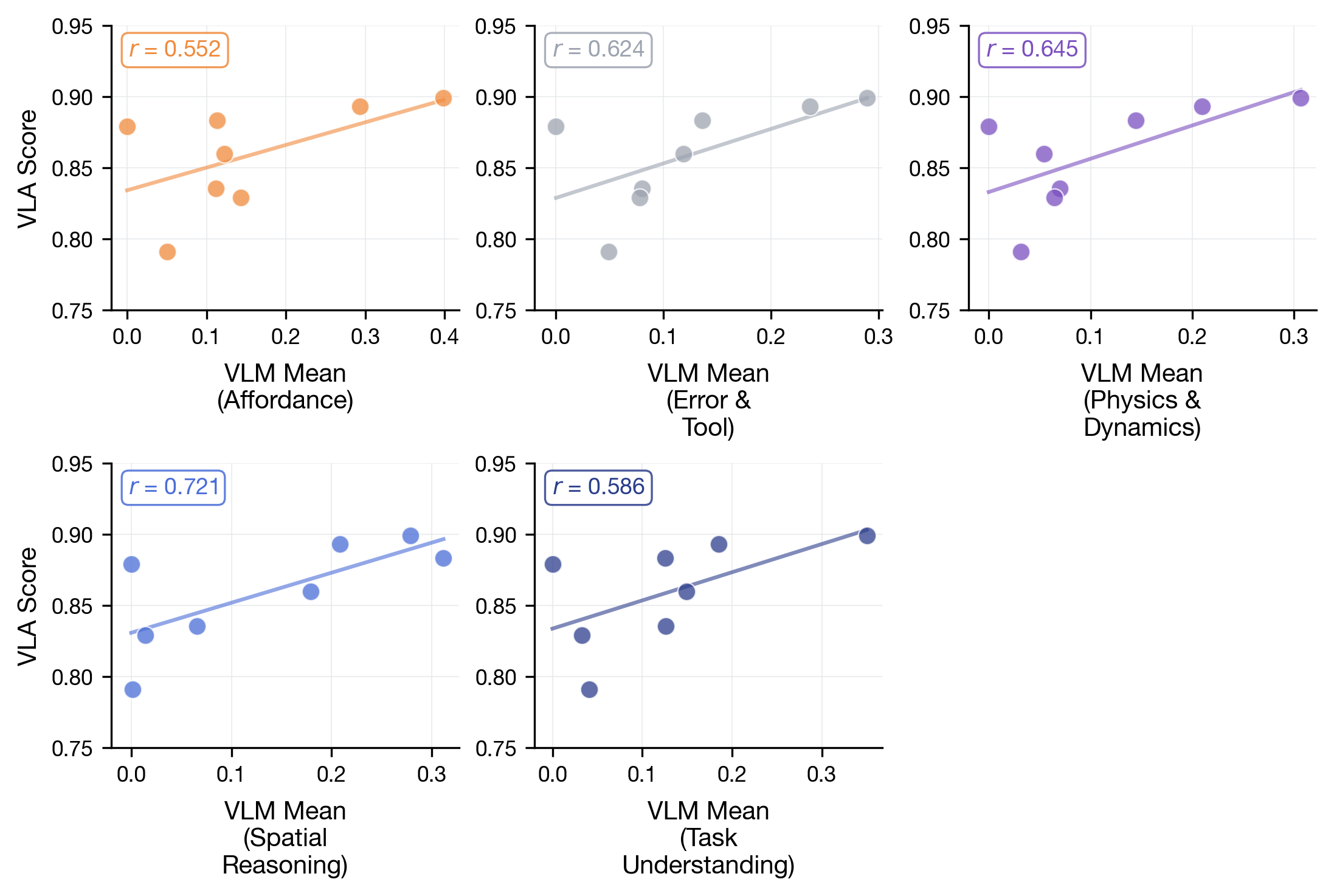}
\caption{\textbf{Per-domain scatter plots between VLM performance on CometBench and VLA success rate on RoboTwin 2.0}}
\label{fig:per_corr}
\end{figure*}

We further provide per-domain scatter plots between VLM performance on CometBench and VLA success rate on RoboTwin 2.0, as shown in~\cref{fig:per_corr}. Each subplot includes a linear regression fit and the corresponding Pearson correlation coefficient. Across all domains, linear fits consistently show positive slopes, reinforcing the conclusion that VLM improvements reliably translate to VLA gains, albeit with domain-dependent strength.

\section{Ablation on the GAP Token}
\label{app:gap}

We use the final checkpoint of our model trained on the full co-training mixture, and extract token-level features on a held-out subset of AgiBot-World-Beta. For every sampled frame, we run a single forward pass with the original training-time data pipeline and record both GAP token and VLM mean pooling baseline representations. The GAP token is taken before being passed into the cross-attention of the DiT-based action expert. The VLM mean pooling baseline is computed by averaging the hidden states of all language and visual tokens at the final layer of the VLM backbone, excluding padding tokens. Both features are extracted in \texttt{bfloat16}. We collect $N{=}3{,}000$ frames sampled uniformly across episodes from the held-out split. To prevent any single trajectory from dominating, we cap the number of sampled frames per episode. 

\paragraph{Visualization Experiment}
AgiBot-World-Beta provides language instructions at the granularity of individual trajectories. We aggregate them into coarse task clusters via standard TF-IDF + KMeans (1--2 grams, English stop-words removed, $k{=}15$). Each resulting cluster is named after the two terms with highest centroid weight (e.g., ``open / drawer'', ``pick / bottle''). We retain frames belonging to the remaining clusters for all subsequent analysis. We use scikit-learn's \texttt{TSNE} with \texttt{perplexity}$=30$, \texttt{init=pca}, \texttt{learning\_rate=auto}, \texttt{max\_iter}$=1000$ and \texttt{random\_state}$=42$.

\paragraph{Action Linear Probing Experiment}
We test whether the GAP token fulfills its role as a physical commonsense aggregation bottleneck. Linear probing measures how much of a target attribute remains linearly separable in this compressed representation. Unlike downstream policy evaluation, probing isolates representation quality from the action head and optimizer. Results show the GAP token aggregates action-relevant commonsense; attributes like gripper state and motion magnitude should be linearly extractable. Probing turns this architectural hypothesis into a quantitative test.

Let $a_t\!\in\!\mathbb{R}^{32}$ denote the action vector predicted at step $t$ inside an action chunk, where a chunk is the contiguous block of $30$ future control commands $\{a_t\}_{t=0}^{29}$ that the policy emits in one forward pass; within each $a_t$, dimensions $6$ and $13$ encode the binary open/close states of the left and right grippers, dimensions $0$--$2$ encode the left end-effector Cartesian $xyz$ position in the robot base frame, and the trailing $14$ dimensions are unused zero padding that we therefore exclude from all probing analysis. From the chunk $\{a_t\}_{t=0}^{29}$ we derive three complementary classification targets that together cover the principal facets the action expert must condition on. 

The first target is the \textbf{gripper state} with four classes defined as: 
\begin{equation}
    y_g = 2\cdot\mathbb{1}[a_0^{(6)}\!>\!0.5] + \mathbb{1}[a_0^{(13)}\!>\!0.5],
\end{equation}
which jointly encodes whether each of the two grippers is open or closed at the start of the chunk and thus probes how well a representation captures the discrete bi-manual control intent that gates every manipulation primitive. 

The second target is the \textbf{motion direction} with six classes, constructed by first computing the displacement of the left end-effector across the chunk $\Delta = a_{29}^{(0:3)} - a_0^{(0:3)}$, then identifying its dominant translation axis $j^{*}=\arg\max_j |\Delta_j|$, and finally assigning $y_d = 2 j^{*} + \mathbb{1}[\Delta_{j^{*}}\!>\!0]$ so that the chunk is labeled by one of $\{\pm x,\pm y,\pm z\}$, which probes whether the representation linearly encodes the geometric direction of the intended motion in workspace coordinates. 

The third target is the \textbf{motion magnitude} with three classes 
$y_m\in\{\text{static, slow, fast}\}$, obtained by binning $\|\Delta\|_2$ at the empirical $1/3$ and $2/3$ quantiles of the dataset, which measures how readily the overall scale of motion, ranging from near stationary holds to fast transport. 

Taken together, these three targets respectively probe discrete control intent, geometric direction, and motion scale, giving a quantitative complement to the qualitative t-distributed Stochastic Neighbor Embedding (t-SNE) visualizations.


\end{document}